\documentclass{article}
\usepackage{epsfig,latexsym,natbib,url,amsmath, amsfonts, eurosym,float}
\begin{document}

\title{\bf Comparing Single and Multiobjective Evolutionary Approaches to the Inventory and Transportation Problem}

\author{ Anna I Esparcia-Alc\'{a}zar\thanks{A.I. Esparcia, A. I. Mart\'{i}nez, E. Alfaro and K. Sharman are with Instituto Tecnol\'ogico de
Inform\'atica, Universidad Polit\'ecnica de Valencia, Spain. E-mail:
\texttt{anna@iti.upv.es}.}, J.J. Merelo\thanks{J.J. Merelo and P.
Garc\'{i}a belong to the Deptartamento de Arquitectura y
Tecnolog\'{i}a de Computadores, Universidad de Granada, Spain.
E-mail:
\texttt{jmerelo@geneura.ugr.es}},\\
 Ana\'{i}s Mart\'{i}nez-Garc\'{i}a,  Pablo
 Garc\'{i}a-S\'{a}nchez\\
Eva Alfaro-Cid and  Ken Sharman}
\date{August 2009\\ {\footnotesize Draft submitted to \emph{Evolutionary
Computation}}}
 \maketitle

\floatstyle{ruled}
\newfloat{Algorithm}{H}{los} %[section]

\hyphenation{accor-ding-ly ana-ly-sis as-so-cia-ted ba-lan-ce ca-te-go-ry
ca-te-go-ries cha-rac-te-ris-tic cha-rac-te-ris-tics com-ple-xi-ty
con-fi-gu-ra-tion con-fi-gu-ra-tions cross-over de-li-ve-ry
di-ffe-ren-ce di-ffe-rent du-ring equi-va-lent ge-ne-ral
ge-ne-ra-li-sa-tion ge-ne-ra-tion ge-ne-ra-tions glo-ba-lly geo-gra-phi-cal gua-ran-te-ed in-di-vi-dual
in-di-vi-duals in-te-rac-tions in-te-rac-tion le-vels li-te-ra-tu-re
ma-trix ma-na-ge-ment me-tho-do-lo-gy mi-ni-mi-se mi-ni-mi-ses mi-ni-mi-sing
mo-du-les par-ti-cu-lar pro-blem pro-blems pro-duct pre-fe-ren-ce
po-pu-la-tion po-pu-la-tions pro-du-cing re-pre-sent re-pre-sents
res-tric-tions rou-ting sa-vings spa-cing ten-ding}

\begin{abstract}
EVITA, standing for \textbf{Ev}olutionary \textbf{I}nventory and
\textbf{T}ransportation \textbf{A}l-gorithm, is a two-level
methodology designed to address the Inventory and Transportation
Problem (ITP) in retail chains. The top level uses an evolutionary
algorithm to obtain delivery patterns for each shop on a weekly
basis so as to minimise the inventory costs, while the bottom level
solves the Vehicle Routing Problem (VRP) for every day in order to
obtain the minimum transport costs associated to a particular set of
patterns.

The aim of this paper is to investigate whether a multiobjective
approach to this problem can yield any advantage over the previously
used single objective approach. The analysis performed allows us to
conclude that this is not the case and that the single objective
approach is in general preferable for the ITP in the case studied. A
further conclusion is that it is useful to employ a classical
algorithm such as Clarke \& Wright's as the seed for other
metaheuristics like local search or tabu search in order to provide
good results for the Vehicle Routing Problem.
\end{abstract}
%\begin{keywords}Inventory and Transportation Problem, Vehicle Routing Problem (VRP), Evolutionary Algorithms, Ant Colony Optimisation, Tabu Search, Clarke and Wright Algorithm.
%\end{keywords}

\section{Introduction}\label{sec:intro}
Given a retail chain and a central depot that supplies it, both
belonging to the same company, we define the Inventory and
Transportation Problem as that whose objective is to minimise the
costs of both inventory and transportation, subject to a number of
constraints imposed at the shop level.

In previous work \citep{esllucarshaand2006b,esllucarshaand2006,esllucarshamer2007a,esllucarshaand2007a,EVITAchapter2008} we employed a single objective approach to this
problem, which aimed at minimising the total weekly cost calculated
as the sum of the inventory and transportation costs. The purpose of the current work is to evaluate the convenience of adopting a multiobjective approach to the problem. Although it is
true that the aim of the ITP is to minimise the global cost and this
is calculated by simply adding the costs of inventory and transport
(both measured in currency units), it is no less true that the
minimisation of both costs are contradictory objectives since, as
was stated above, reducing the cost of transport implies increasing
the cost of inventory and vice versa. For this reason it is possible
that a multiobjective approach will obtain better results in this
problem, and this is what we set out to investigate.

With this aim in mind we have carried out an extensive series of
experiments using the eight instances used in
\citep{EVITAchapter2008}, plus two new ones\footnote{Nine instances are taken from
\url{http://branchandcut.org/VRP/data/} and the remaining one from
\url{http://www.fernuni-hagen.de/WINF/touren/inhalte/probinst.htm}}. The
set of restrictions, consisting of the characteristics of the
vehicles employed and the working hours of the drivers (see Table
\ref{tab:vehicledata}), plus the parameter configuration of the
evolutionary algorithm are kept as in that work.

 In the multiobjective approach we have used the NSGA-II algorithm
 \citep{nsgaII2000}, which is described in more detail in Appendix I.

A further objective is related to the fact that the choice of the
algorithm that yields the transportation routes (the \emph{VRP
solver}) can play a significant part in the performance of the whole
algorithm \citep{esllucarshaand2006b}. Here we will employ improved versions of three different
algorithms that have been employed in the literature for solving the
VRP and will compare them on the selected problem instances, with the aim
to determine whether there is a relation between the instance
characteristics %(such as size and eccentricity)
and the performance of the VRP solver. The VRP algorithms studied are improved variants of the three that
obtained best results in \citep{EVITAchapter2008}: tabu search
\citep{cogenla97}, ant colony optimisation \citep{dongxiang2006} and
a classical VRP solving technique,  Clarke and Wright's algorithm
\citep{clawri64}.

The rest of the paper is structured as follows. Section
\ref{sec:background} provides background on the problem; it contains
a summary of the state of the art in this and related problems and a
detailed description of the ITP. The top and lower levels of the
algorithm are described in Sections \ref{sec:top-level} and
\ref{sec:lower_level}, the latter containing the details of the
three VRP solvers employed. The experimental setup is given in
Section \ref{sec:experiments}, with results and analysis thereof
contained in Section \ref{sec:results}. Finally, Section
\ref{sec:conclusions} presents the conclusions and outlines future
areas of research.

\section{Problem background} \label{sec:background}

The ITP deals with the management of two different aspects of retail
chain logistics, inventory costs and transportation costs, which
have both received lots of attention in the logistics literature.
Here we will describe how they have been addressed in the past and
what aspects are particularly relevant to our case.

\subsection{The Vehicle Routing Problem}\label{subsec:vrp}

The Vehicle Routing Problem (VRP) consists on finding an optimal set
of delivery routes from a depot to a set of customers to serve
\citep{tovi01}. The routes must start and finish in the depot and
each customer must be served by one and only one vehicle\footnote{No
single customer can consume more than the capacity of any one
vehicle.}, which means a customer cannot be contained in more than
one route. Different versions of the problem have slightly different
objectives or ways to define optimality: it can refer to finding the
minimum cost, employing minimum time, minimum number of delivery
vehicles or combinations of these and other factors.

Amongst these variants of the problem the most popular is the
\emph{capacitated} VRP, or CVRP, which refers to the fact that each
delivery vehicle has a limited capacity\footnote{A VRP in which
vehicles had infinite capacity could be subsumed under the general
Traveling Salesman Problem}. Also popular is the VRP with \emph{time
windows}, or VRPTW, in which each customer must be served during a
specified time interval or window.

Another variant of the VRP relevant to our problem is the
\emph{periodic} vehicle routing problem (PVRP), which appears when
customers have established a predetermined delivery frequency and a
combination of admissible delivery days within the planning horizon.
The objective is to minimise the total duration of the routes, while
the restrictions usually involve a limited capacity of the delivery
vehicles and a maximum duration of each itinerary. See for instance,
\citep{cogenla97} and \citep{tovi01} for a general description of
the periodic VRP.

In this paper we are concerned with the CVRP, simply referred to as
VRP. The only limitation will be in the capacity of the vehicles
used, and not in their number. We will also consider that the fleet
is homogeneous, i.e. only one type of vehicle is used, with unique
values of average velocity and capacity.

For simplicity reasons we will consider that the customers (shops in
the retail chain in our case) have no time windows, i.e. the
deliveries can take place at any time. However, the hours a driver
can work are limited by regulations and this has to be taken into
account in the time needed per delivery, as is the unloading time.
Also, it should be possible to serve all customers; this means that
routes from the depot to any customer and back must take less than
the working hours of the driver.

Finally, we will consider that the transport cost only includes the
cost per kilometer; there is no cost attached to either the time of
use of the vehicle or to the number of units delivered, nor there is
a fixed cost per vehicle.

%\section{State of the art}\label{sec:state_of _art}
\subsection{Inventory and transportation
management}\label{subsec:inv_trans_management} One step beyond the
VRP is the problem of optimising simultaneously the costs of
inventory and transportation. In \citep{esllucarshamer2007a} the
reader can find a review of different approaches employed in the
literature.

Amongst these, one of the most relevant to our case is the inventory
routing problem (IRP), which arises when a vendor delivers a single
product and implements a Vendor Inventory Management (VIM)
\citep{cetinkaya:vmi} policy with its clients, so that the vendor
decides the delivery (time and quantity) in order to prevent the
clients from running out of stock while minimising transportation
and inventory holding costs \citep{camsav04}. However, retail chains
cannot be addressed in this way, as thousands of items are involved.

The work presented here focuses on the Inventory and Transportation
Problem (ITP), which was first described in \citep{manolojosepe}
with the aim of addressing the case of retail chains. Thus, the ITP
can be viewed as a generalisation of the IRP to the multiproduct
case. Additionally, it can also be viewed as a variant of the PVRP
described previously that includes inventory costs\footnote{I.e. the
PVRP can be seen as an ITP in which the inventory costs are zero.}
and a set of delivery frequencies instead of a unique delivery
frequency for each shop.

%\subsection{The Inventory and Transportation Problem} \label{subsec:ITPdescr}
%The Inventory and Transportation Problem (ITP) arises when the objective is to minimise both the
%transport and inventory costs of a retail chain supplied from a central depot, both owned by the same company and subject to the
%operational constraints taking place at the shop level; for instance, just a few \emph{delivery frequencies} are allowed for each shop.

The main feature that differentiates the ITP %this problem
from other similar supply chain management ones addressed in the
literature
\citep{cetinkaya:vmi,dosSantos,Federgruen:CVRIA,Sindhuchao:IRS,Viswanathan}
is that we have to decide on the frequency of delivery to each shop,
which determines the size of the deliveries. The inventory costs can
then be calculated accordingly, assuming a commonly-employed
periodic review stock policy for the retail chain shops. Besides,
for a given delivery frequency, expressed in terms of number of days
a week, there can also be a number of \emph{delivery patterns}, i.e.
the specific days of the week in which the shop is served. Once
these are established, the transportation costs can be calculated by
solving the VRP for each day of the week.

Because a pattern assumes a given frequency, the problem is limited
to obtaining the optimal patterns (one per shop) and set of routes
(one set per day). The optimum is defined as a combination of
patterns and routes that minimises the total cost, which is
calculated as the sum of the individual inventory costs per shop
(inventory cost) plus the sum of the transportation costs for all
days of the week (transport cost). These two objectives are in
general contradictory: the higher the frequency of delivery the
lower the inventory cost, but conversely, a higher frequency
involves higher transportation costs.

The operational constraints at the shop level are imposed by the
business logic and can be listed as follows
\citep{EVITAchapter2008}:
\begin{enumerate}

\item{A periodic review stock policy is applied for the shop items. %So, a target cycle service level has been established for every item category, which is usually lower for the slow-moving ones.
This means that the decision of whether to deliver to a particular
shop is taken centrally and not at the shop. As a consequence,
stockout is allowed.}\label{it:periodic_review}

\item{Shops have a limited stock capacity.}\label{it:limited_stock_capacity}

\item{The retail chain tries to fulfil backorders in as few days as possible, so
there is a lower bound for delivery frequency, which depends on the
target client service level.}\label{it:lower_bound_on_delivery_freq}

\item {The expected stock reduction between replenishments (deliveries) cannot be too high in order to avoid
two problems: (a) the unappealing empty-shelves aspect of the shop
just before the replenishment; and (b) replenishment orders too
large to be placed on the shelves by the shop personnel in a short
time compatible with their primary selling
activity.}\label{it:stock_reduction_not_too_high}

\item{Conversely, the expected stock reduction must
be high enough to perform an efficient allocation of the
replenishment order.}\label{it:stock_reduction_not_too_low}

\item{As a consequence of points \ref{it:limited_stock_capacity} to \ref{it:stock_reduction_not_too_low} above, not all frequencies are admissible for all
shops; in general, very high and very low frequencies are
undesirable. For instance, frequency 1 (one delivery per week) is
not applicable to most shops.}% for various reasons. On the one hand, many shops do not have the storage capacity required for the big deliveries imposed by such a low delivery frequency. On the other hand, big deliveries imply that the shop staff must devote a long time to the
%management of the stock, a time that would be more profitably employed in selling.}

 \item{Sales are not uniformly distributed over the time horizon (week),
     tending to increase over the weekend. Hence, in order to match
     deliveries to sales,
only a given number of \emph{delivery patterns} are allowed for
every feasible frequency. For instance, a frequency-2 pattern such
as (Mon, Fri) is admissible, while another of the same frequency
such as (Mon, Tues) is not.}
\item{Although we are dealing with thousands of items, the load
is containerised; hence, the size of the deliveries is expressed as
an integer, representing the number of roll-containers.}
\end{enumerate}

Thus, to summarise, our task involves finding:
\begin{itemize}
\item{The \emph{optimal set of patterns}, $\mathcal{P}_{opt}$,
with which all shops can be served. A pattern $p$ represents a set
of days in which a shop is served which implies a delivery frequency
(expressed as number of days a week) for the shop}
\item{The \emph{optimal routes} for each day of the working week, by
solving the VRP for the shops allocated to that day by the
corresponding pattern}
\end{itemize}

\subsection{Objective functions in EVITA}\label{subsec:evita}

EVITA operates in two levels: the lower one deals exclusively with
the transportation costs per day and the top one incorporates these
and the inventory costs into the total costs.

In the single objective case, the optimum is defined as the solution
that minimises the total cost, given by the function
\begin{equation}\label{eq:fitness1}
f = TotalCost = InventoryCost + TransportCost
\end{equation}

In the multiple objective approach, we will deal with two separate
cost functions:

\begin{eqnarray}\label{eq:fitness2}
f_i = InventoryCost \\
f_t = TransportCost
\end{eqnarray}

However, we will still use the total cost defined for the single objective case to compare the results between multi and monoobjective solutions. The reason is that the company is interested in spending less, irrespective of where the reduction comes from, and, at the end of the day, both costs come in euros.

Inventory costs are computed from the patterns for each shop by
taking into account the associated delivery frequency and looking up
the inventory cost per shop in the corresponding table. An example
of the latter is given in Table \ref{tab:inventorycost}.

\begin{table*}[htb]
\begin{center}\setlength{\tabcolsep}{0.7mm}\caption{\label{tab:inventorycost}Inventory cost (in Euro) and size of the deliveries per shop (expressed in roll containers) depending on the delivery frequency (in days). Missing data corresponds to
frequencies that are not admissible for each shop.}% Shop number 0 corresponds to the depot. }
%{\scriptsize
\begin{tabular} {|c|ccccc|ccccc|}
\multicolumn{11}{c}{ }\\
\hline
    &   \multicolumn{5}{c|}{ }   &    \multicolumn{5}{c|}{ }    \\
    &   \multicolumn{5}{c|}{Inventory cost }   &    \multicolumn{5}{c|}{Delivery size}    \\
  &    \multicolumn{5}{c|}{(\euro)}           &   \multicolumn{5}{c|}{{\scriptsize(roll containers)}}  \\%\cline{2-11}
 &    \multicolumn{5}{c|}{}                                   &   \multicolumn{5}{c|}{}\\
 Shop  \#& \multicolumn{5}{c|}{{\scriptsize Frequency (days)} } & \multicolumn{5}{c|}{{\scriptsize Frequency (days)} }\\
%\hline \\
  &   1   &   2   &   3   &   4   &   5   &   1   &   2   &   3   &   4   &   5      \\
\hline
1   &   - &   - &   - &   336 &   325 &   -   &   -   &   -   &   2   &   2      \\
2   &   - &   - &   - &   335 &   325 &   -   &   -   &   -   &   2   &   2  \\
 $\cdots$   &   $\cdots$ &    $\cdots$ &    $\cdots$ &    $\cdots$ &    $\cdots$ &    $\cdots$   &    $\cdots$   &    $\cdots$   &    $\cdots$   &  \\
N  &   - &   311 &   293 &   286 &   284 &   -   &   3   &   2   &   2   &   1   \\
\hline \multicolumn{11}{c}{ }\\
\end{tabular} %}
\end{center}
\end{table*}

For instance, let us assume that shop N was assigned a pattern of
frequency 4; we would look up in the table the inventory cost for
the shop at that frequency, which is 286\euro. Proceeding in the
same way with all shops and adding up the results we would obtain
the total inventory cost.

Transport costs are obtained by solving the VRP with one of the
algorithms under study. The demands (size of deliveries) of each
shop would also be taken from Table \ref{tab:inventorycost}. In the
example above, for shop N at frequency 4 the delivery size is 2
roll containers.

Problem data is freely available from our group website:
\url{http://casnew.iti.es/}\footnote{Its use is
  subject to the condition that this or other papers on the same
  subject by the authors are mentioned}

\section{The top level: Evolutionary Algorithm}\label{sec:top-level}

The top level in EVITA is an evolutionary algorithm in which a
population of individuals (candidate solutions) undergoes evolution
following Darwinian principles. Each individual is a set of patterns
$\mathcal{P}$ represented as a vector of length equal to the number
of shops to serve ($nShops$),

\[ P = ( p_1, p_2, \cdots, p_{nShops}) \]

and whose components $p_i, ~1 \leq p_i \leq 2^d -1$, $ i \in \{1 \dots
nShops\}$ are integers
representing a particular delivery pattern, where $d$ is the number of days in the working week.

The relationship between patterns and delivery days is made at bit
level. Each pattern is formed by $d$ bits and each day corresponds
to one bit: 1 means that the store is visited that day and 0 that it
is not. In our case the working week has 5 days (Monday to Friday)
so $d=5$. Hence patterns are coded by the rightmost 5 bits of the
integer value. For instance pattern 21, i.e. 10101 in binary,
corresponds to deliveries on Monday, Wednesday and Friday.

The total number of possible patterns is $2^d -1$; in our case
$2^5-1= 31$, so patterns range from 1 to 31 or, alternatively, from
00001 to 11111 (obviously pattern 00000, i.e. not delivering any
day, is not admissible). However, as was stated in
Subsection \ref{subsec:inv_trans_management}, not all patterns are suitable for all shops. Hence, $p_i$
must be contained in the set of admissible patterns, $p_i \in$
$\mathcal{P}$$_{adm}$. The elements in $\mathcal{P}$$_{adm}$ are
given in Table \ref{tab:patterns}.

\begin{table}[hbt]
\begin{center}\setlength{\tabcolsep}{2mm}\caption{\label{tab:patterns}The
admissible patterns and their respective frequencies. The 1
represents that the shop is served on that day, the 0 that it is
not. As a consequence of the business logic, we will only consider
11 patterns out of the 31 that are possible.}
{\footnotesize
%{\scriptsize
\begin{tabular}  {ccccccc}  \multicolumn{7}{c}{}\\\hline
\textbf{Pattern Id.}& \textbf{Frequency} (days) &\textbf{Mon}   & \textbf{Tues}  & \textbf{Wed}  & \textbf{Thurs}  & \textbf{Fri} \\
\hline
5 &2& 0 &0 & 1 & 0 & 1  \\
9 &2& 0 & 1  &0 & 0& 1 \\
10 &2& 0 & 1 &0 &  1  &0\\
11 &3& 0 & 1  &0 &  1  & 1  \\
13 &3& 0 & 1  & 1 & 0 & 1 \\
17 &2&  1  &0 &0& 0 & 1  \\
18 &2&  1  &0 &0 &  1 & 0 \\
21 &3&  1  &0 & 1  & 0 & 1  \\
23&4&  1  &0 & 1  &  1  & 1  \\
29&4&  1  & 1  & 1  & 0 & 1  \\
31&5&  1  & 1  & 1  &  1  & 1 \\
\hline \multicolumn{7}{c}{} \\
\end{tabular}} \end{center}
\end{table}

A point to note is that although the binary representation of the
patterns is convenient in order to figure out what delivery days are
associated to a pattern and also for calculating its corresponding
frequency, in the genetic algorithm we will be considering the
integer values only.

%\begin{figure*}[bt]
%\begin{center}
%\resizebox{13.5cm}{!}{
%\includegraphics{figures/flowchartEC.eps}
%} \caption{Flow chart of the evolutionary algorithm employed at the
%top level for the single objective case.} \label{fig:flowchart}
%\end{center}
%\end{figure*}

%The flow diagram for the top level evolutionary algorithm is depicted in Figure \ref{fig:flowchart} and
The details of the top level evolutionary algorithm are given in
Table \ref{tab:evopars-toplevel}. The pseudo-code for the evaluation
function is given in Algorithm \ref{evaluation-pseudo-code}.

\begin{Algorithm}[tb]
\begin{center}\setlength{\tabcolsep}{2mm}
\renewcommand{\arraystretch}{1.2}
{\footnotesize \textsf{
\begin{tabular}{l} \multicolumn{1}{c}{}\\ %\hline
\textbf{Procedure} Evaluate \\
\textbf{input:} \\
\qquad Chromosome  $\{p_1, \dots, p_{nShops}\}$, \\
\qquad problem data tables \\
\qquad $costPerKm$ \\
\textbf{output:} Fitness \\
\qquad  [Calculate inventory cost] \\
\qquad \quad InventoryCost = 0 \\
\qquad \quad \textbf{for} $i=1$ \textbf{to} $nShops$ \\
\qquad \quad \quad Look up frequency $f_i$ for pattern $p_i$ \\
\qquad \quad \quad Look up cost $c_i$ for shop $i$ and frequency $f_i$ \\
\qquad \quad \quad $InventoryCost += c_i$ \\
\qquad  [Calculate transportation cost] \\
\qquad \quad $day = Monday$;\\
\qquad \quad  $totalDistance = 0$\\
\qquad \quad\textbf{repeat}\\
\qquad \quad \quad Identify shops to be served on $day$\\
\qquad \quad \quad Run \emph{VRP solver} to get $dayDistance$\\
\qquad \quad \quad $totalDistance += dayDistance$\\
\qquad \quad \quad $day++$\\
\qquad \quad\textbf{until} $day = lastWorkingDay$; \\
\qquad \quad $TransportCost = totalDistance * costPerKm$ \\
\qquad  \textbf{if} \emph{multiobjective} \\
\qquad \quad \textbf{return} $InventoryCost,~ TransportCost$\\
\qquad  [Calculate total cost] \\
\qquad \quad $TotalCost = InventoryCost + TransportCost$\\
\qquad    \textbf{return} $TotalCost$\\
\textbf{end procedure;}  \\
%\hline
\end{tabular}
}}\caption{\label{evaluation-pseudo-code}Evaluation
function.}\end{center}
\end{Algorithm}

\begin{table*}[tb]
\begin{center}\setlength{\tabcolsep}{2mm}
\renewcommand{\arraystretch}{1.3}
\caption{\label{tab:evopars-toplevel}Configuration of the top-level
evolutionary algorithm employed (both mono and multiobjective).}
%{\scriptsize
%{\footnotesize
\begin{tabular}{lp{8.5cm}} \multicolumn{2}{c}{}\\\hline
\textbf{Encoding} & The gene $i$ represents the pattern for shop
$i$.\\
&The chromosome length is equal to the number of shops
($nShops$).\\
 \textbf{Selection} & Tournament
in 2 steps. To select each parent, we take $tSize$ individuals
chosen randomly and select the best.\\
& For the single objective algorithm the best 10 individuals of each
generation are preserved as the
elite.\\
\textbf{Evolutionary operators} & 2 point crossover and 1-point
mutation. \\
& The mutation operator changes the pattern for
1 shop in the chromosome.  \\
 \textbf{Termination criterion}& Terminate when the total number of generations (including the initial one) equals 100.\\
\textbf{Fixed parameters} & Population size, $popSize =100$\\
& Tournament size, $tSize = 2$\\
& Mutation probability, $pM = 0.2$ \\
& Crossover probability, $pC = 1$ \\\hline
\multicolumn{2}{c}{}\\
\end{tabular}%}
\end{center}
\end{table*}

\section{Lower level: Solving the VRP}\label{sec:lower_level}

%Four
The calculation of the fitness of an individual requires an
algorithm to solve the VRP (a \emph{VRPsolver}). In this work three
algorithms have been tested for this purpose, namely:\begin{itemize}
\item[ ] \textbf{CWLS}, the classical Clarke and
Wright's algorithm \citep{clawri64} enhanced with local search,
which is presented in subsection \ref{subsec:CW}
\item[ ] \textbf{ACO}, a bioinspired ant colony optimisation
algorithm \citep{dostu2004}, described in subsection
\ref{subsec:ACO}, and
\item[ ] \textbf{CWTS}, tabu search \citep{gloco02} seeded with a solution obtained by Clarke and Wright algorithm,
as described in subsection \ref{subsec:TS}
%\item[ ]\textbf{EC}, another evolutionary algorithm \citep{bafomi2000}, described in subsection \ref{subsec:EC}
\end{itemize}

In \citep{EVITAchapter2008} we also employed a fourth algorithm as
\emph{VRPsolver}, evolutionary computation. However, the results
obtained there were not very promising, which is why it has been
omitted in this study.

\subsection{Clarke and Wright's algorithm}\label{subsec:CW}
Clarke and Wright's algorithm \citep{clawri64} is based on the
concept of \emph{saving}, which is the reduction in the traveled
length achieved when combining two routes. We employed the parallel
version of the algorithm, which works with all routes
simultaneously.

Due to the fact that the solutions generated by the C\&W algorithm
are not guaranteed to be locally optimal with respect to simple
neighbourhood definitions, it is almost always profitable to apply
local search to attempt and improve each constructed solution. For
this purpose we designed a simple and fast local search method,
which consists on performing 2-interchanges on the  solution
obtained by the C\&W algorithm. Every possible pair of shops is
exchanged, first between shops in the same route and then between
shops in different routes. If at any time an invalid route is generated
(because the restrictions on time or capacity are violated) the
depot is inserted where required in the route. The best neighbour
solution will be the one with a lower associated transport cost.

This combination of C\&W's algorithm with local search is what we
have termed CWLS, the pseudo-code for both can be found in
Algorithms \ref{CWLS-code} and \ref{alg:localSearch}.

\subsection{Ant Colony Optimisation}\label{subsec:ACO}

Some ant systems have been applied to the VRP (see for instance
\citep{colriz07,gelapo02}) with various degrees of success. Ant
algorithms are derived from the observation of the self-organized
behavior of real ants \citep{dostu2004}. The main idea is that
artificial agents can imitate this behavior and collaborate to solve
computational problems by using different aspects of ants' behavior.
One of the most successful examples of ant algorithms is known as
``ant colony optimisation'', or ACO, which is based on the use of
pheromones, chemical products dropped by ants when they are moving.

Each artificial ant builds a solution by choosing probabilistically
the next node to move to among those it has not visited yet. The choice is biased by pheromone
trails previously deposited on the graph by other ants and some
heuristic function. Also, each ant is given a limited form of memory
in which it can store the partial path it has followed so far, as
well as the cost of the links it has traversed. This, together with
deterministic backward moves, helps avoiding the formation of loops
\citep{dostu2004}.

In this work we employed a variant of ACO described by \cite{xiqiyoda2006} which differs from the original ACO
algorithm in three aspects: (1) the way the pheromone matrix is
updated, (2) the transition function and (3) that
$\lambda$-interchanges are used instead of local search.

Two ways of updating the pheromone matrix $\tau$ are defined:
\textbf{local updating} and \textbf{\emph{a posteriori} updating}
(i.e. taking place after all ants have built their solutions). The
former consists of adding $1/d_{i,j}$ to each element $\tau_{i,j}$
of the pheromone matrix, with $d_{i,j}$ being the distance between
shops $i$ and $j$. The latter is given by Equation
\ref{eq_UpdatePosteriori}. Here the best path built in iteration $t$
receives a reinforcement while the worst path is reset to the
initial pheromone value, $\tau_0$.

\begin{equation}\label{eq_UpdatePosteriori}
\tau_{i,j} = \left\{
  \begin{array}{ll} \setlength{\tabcolsep}{7mm}
     \tau_{i,j} (\rho + \frac{1-\rho}{minCost_t}) & \parbox[c][1.2cm][c]{7.5cm}{if $(i,j$) is an arc visited by
    the \emph{best} ant\\ in iteration $t$}\\
    \tau_0 & \parbox[c][1.2cm][c]{7.5cm}{if $(i,j)$ is an arc visited by
    the \emph{worst} ant \\in iteration $t$}\\
  \end{array}
\right.
\end{equation}

where $minCost_t$ is the minimum cost obtained in iteration $t$ and
the value of $\rho$ is automatically corrected in each iteration, as
follows,

\begin{equation}\label{eq:UpdateRho}
\rho_{t} = \left\{
  \begin{array}{ll}
    0.95 \rho_{t-1} & \hbox{if } 0.95\rho_{t-1} < \rho_{min} \\
    \rho_{min} & \hbox{otherwise}\\
  \end{array}
\right.
\end{equation}

The second difference with respect to the original ACO is the
transition function. In our ACO an ant located at shop $i$ will
select as its next shop  $j$ the one given by Equation
\ref{eq_transition}, with probability $p_t$,

\begin{equation}\label{eq_transition}
j =argmax_{S}
[\tau_{i,j}]^\alpha[\eta_{i,j}]^\beta[\mu_{i,j}]^\gamma
\end{equation}

where  $S$ is the list of shops not visited yet, $\tau$ is the
pheromone matrix, $\eta$ is the heuristic function,
\[\eta_{i,j} = \frac{1}{d_{i,j}}\]

 and finally,

 \[\mu_{i,j}=d_{i,0}+d_{0,j}-d_{i,j}\]

corresponds to the concept of saving used in  Clarke \& Wright's
algorithm, with shop 0 being the depot\footnote{The transition
function used in \citep{xiqiyoda2006} also takes into account the
time window of each customer; we have skipped this because we do not
consider time windows in our problem.}. Alternatively, the next shop
$j$ will be uniformly selected at random from $S$.

The parameters $\alpha$, $\beta$ and $\gamma$ (whose sum does not
necessarily equal 1) measure the relative importance of each
component. The probability value $p_t$ is dynamically adjusted at
runtime in a similar way as for $\rho_t$, following Equation
\ref{eq:pUpdate}.

\begin{equation}\label{eq:pUpdate}
p_{t} = \left\{
  \begin{array}{ll}
    0.95 p_{t-1} & \qquad \hbox{if } \quad 0.95~ p_{t-1} \geq p_{min} \\
    p_{min} & \qquad \hbox{otherwise}\\
  \end{array}
\right.
\end{equation}

%where $p_{Min}$ is used to ensure the achievement of the definite choosing chance even if $p_{t}$ is too small.

Finally, instead of using local search in the closest neighbours as
in conventional ACO, we use $\lambda$-interchanges, a concept
borrowed from the CWLS algorithm described earlier.

The pseudo-code for the ACO algorithm employed here is given in
Algorithm \ref{alg:HybridACO-code}, its transition function is shown
in Algorithm \ref{alg:transition-code} and the parameters used are
given in Table \ref{tab:paramHyACO}.

\begin{table}[bht]
\begin{center}\setlength{\tabcolsep}{2mm}\caption{Parameters for ACO. The values were obtained by trial and error, as explained in \citep{anais2008}.
Notice that $\alpha$, $\beta$ and $\gamma$ do not necessarily add up
to 1.} \label{tab:paramHyACO}
\renewcommand{\arraystretch}{1.2}
%{\footnotesize
\begin{tabular} { l p{7.5cm}  }%{ p{3cm} p{7cm} }
\multicolumn{2}{c}{ }\\
\hline \textbf{Number of Iterations} &   $nIters = 50$ \\
\textbf{Number of Ants} & $nAnts =
25$\\\textbf{Transition function} & %\parbox[c][5cm][c]{7.5cm}{
Probability, $p_t$\\
&\qquad Initial value, $p_0=0.8$\\
&\qquad Minimum value, $p_{min}= 0.1$\\
&Weights: \\
&\qquad Pheromone: $\alpha=0.2$\\
&\qquad Heuristics: $\beta = 0.8$\\
&\qquad Savings: $\gamma = 0.3$\\ %}\\
 \textbf{Pheromones } & Initial value, $\tau_0 = 0.5$\\
 & Update factor, $\rho$\\
 & \qquad Initial value,  $\rho_{0}=1$ \\
 & \qquad Minimum value, $\rho_{min} = 0.1$\\
\hline
\end{tabular}%}
\end{center}
\end{table}

\subsection{Tabu Search}\label{subsec:TS}
Tabu Search (TS) is a metaheuristic introduced by Glover and
Kochenberger in order to allow Local Search (LS) methods to overcome
local optima \citep{gloco02}. The basic principle of TS is to pursue
LS whenever a local optimum is found by allowing non-improving
moves; cycling back to previously visited solutions is prevented by
the use of memories, called \emph{tabu lists} \citep{gen2002} which
last for a period given by their \emph{tabu tenure}. The main TS
loop is given in Algorithm \ref{alg:TS-code}.

To obtain the best neighbour of the current solution we must move in
the solution's neighbourhood, avoiding moving into older solutions
and returning the best of all new solutions. This solution may be
worse than the current solution. For each solution we must generate
all possible and valid neighbours whose generating moves are not
tabu. If a new best neighbour is created, the movement is inserted
into the Tabu List with the maximum tenure. This movement is kept in
the list until its tenure is over. The tenure can be a fixed or
variable number of iterations.

The way to obtain the best neighbour is described in Algorithm
\ref{alg:TS-bestneighbour-code}. Table \ref{tab:paramTS} lists the
relevant information about the algorithm.

In order to improve over the TS algorithm used in
\citep{EVITAchapter2008}, we considered \emph{seeding} the TS with a
good solution. The solution chosen as a starting point was one
obtained by the C\&W algorithm. Statistical analysis carried
out\footnote{Mann-Whitney test on the \emph{relative percentage deviation} (RPD), given by Eqn (\ref{eqn:RPD}), of the best results of 10
runs per algorithm over the 8 instances of groups A and B, see table
\ref{tab:problem_instances}.} shows that the seeded algorithm
obtains significantly better results than when the initial solution
is obtained at random. For this reason, in the rest of this work we
have employed this improved version, which we have termed CWTS.

\begin{table}[bht]
\begin{center}\setlength{\tabcolsep}{2mm}\caption{Configuration for CWTS
algorithm. The parameters were tuned heuristically, after several
test launches. The chosen values (small value for the Tabu Tenure and
large number of iterations without solution improvement) sacrifice
execution time in order to obtain better solutions. }
\label{tab:paramTS}
\renewcommand{\arraystretch}{1.2}
%{\footnotesize
\begin{tabular} { l p{9cm}  }%{ p{3cm} p{7cm} }
\multicolumn{2}{c}{ }\\
\hline
\textbf{Initial solution} & A  solution obtained by Clarke \&
Wright's algorithm\\
\textbf{Possible moves} & Swap shop $i$ with shop $j$, both in same
route \\ & Swap shop $i$ with shop $j$, in different routes \\
& Create new route with shop $i$ only\\
\textbf{Tabu tenure} & 12 iterations\\
 \textbf{Termination criterion} &   20 iterations without improvement \\
\hline
\end{tabular}%}
\end{center}
\end{table}

\section{Experiments}\label{sec:experiments}

This section is devoted to present the data we have used in the
problem (subsection \ref{subsec:data}) and the experimental
procedure we have followed (in the next subsection,
\ref{subsec:exp-procedure}).

\subsection{Problem data} \label{subsec:data}

As explained above, we will employ a number of \emph{geographical
layouts} available on the web. We have selected our instances so as
to achieve the maximum representation on three
categories:\begin{itemize}
\item{\textbf{size}, given by the number of shops,}
\item{\textbf{distribution}. We consider two kinds of distributions:
\emph{uniform} and \emph{in clusters}, corresponding to shops that
are scattered more or less uniformly on the map or grouped in
clusters and,}
\item{\textbf{eccentricity}. This represents the distance
between the depot and the geographical centre of the distribution of
shops. The coordinates of the geographical centre are calculated as
follows: \begin{equation*}
(x_{gc} ,y_{gc}) = \frac{1}{nShops}\sum_{i=1}^{nShops} (x_i,y_i) %\\
%y_{gc} = \sum_{i=1}^{nShops} y_i \\
\end{equation*}An instance with low eccentricity (in practise, less
than 25) would have the depot centered in the middle of the shops
while in another with high eccentricity (above 40) most shops would
be located on one side of the depot.}\end{itemize}

We chose ten instances with different levels of each category, see
Table \ref{tab:problem_instances}.

\begin{table}[hbt]\begin{center}\caption{ \label{tab:problem_instances}Problem instances used in the experiments and their characteristics. The first nine have been taken from \texttt{http://branchandcut.org/VRP/data/} and the last one from \texttt{http://www.fernuni-hagen.de/WINF/touren/inhalte/probinst.htm}.
 }
\setlength{\tabcolsep}{1.5mm}%{\scriptsize
\begin{tabular} {llccc}
\multicolumn{5}{c}{ }\\
\hline
\textbf{ID}  &   \textbf{Instance}    &   \textbf{Distribution }   &   $nShops$    &   \textbf{Eccentricity}    \\
\hline
A32 &   A-n32-k5.vrp    &   uniform  &   31 &   47.4  \\
A33 &   A-n33-k5.vrp    &   uniform  &   32 &   20.2  \\
A69 &   A-n69-k9.vrp    &   uniform  &   68 &   15.3  \\
A80 &   A-n80-k10.vrp   &   uniform  &   79 &   63.4  \\
B35 &   B-n35-k5.vrp    &   clusters &   34 &   60.5  \\
B45 &   B-n45-k5.vrp    &   clusters &   44 &   16.6  \\
B67 &   B-n67-k10.vrp   &   clusters &   66 &   19.9  \\
B68 &   B-n68-k9.vrp    &   clusters &   67 &   49.2  \\
P100&   P-n101-k4.vrp   &   uniform  &  100 &    1.59 \\
X200&     c1\_2\_1.txt    &   clusters &  200 &    8.15 \\
\hline
\multicolumn{5}{c}{ }\\
\end{tabular}%}
\end{center}\end{table}

%In the original problem the $n$ indices represented the number of shops plus one and the $k$ indices the maximum number of vehicles
%allowed. In our case we will ignore the latter data, as we are not considering restrictions in the number of vehicles.

It must be noted that we are only using the spatial location and not
other restrictions given in the bibliography, such as the number of
vehicles  or the shop demand values. As pointed out earlier, a main
characteristic of our problem is that the latter is a function of
the delivery frequency, so we had to use our own values for the
demands.

We also added a list of \emph{admissible patterns}, which are given
in Table \ref{tab:patterns}, and \emph{inventory costs}, an example
of which is given in Table \ref{tab:inventorycost}. The inventory,
demand and admissible patterns data were obtained from Druni SA, a
major regional Spanish drugstore chain.

Finally, we have used the vehicle data given in Table
\ref{tab:vehicledata}.

\begin{table}[bht]
\begin{center}\setlength{\tabcolsep}{2mm}\caption{\label{tab:vehicledata}Data for Vehicle Routing Problem}%{\footnotesize
\begin{tabular} {ll} \multicolumn{2}{c}{} \\   \hline

\textbf{Vehicle capacity}    & 12 roll containers\\
\textbf{Transportation cost} &    0.6 \euro /Km \\ %\hline \hline
\textbf{Average speed}  & 60 km/h \\
\textbf{Unloading time}  & 15 min  \\
\textbf{Maximum working time} &8h\\
\hline
 \multicolumn{2}{c}{ }\\
\end{tabular}%}
\end{center}
\end{table}

\subsection{Experimental procedure}\label{subsec:exp-procedure}

Our aim is, on the one hand, to evaluate whether the multiobjective
approach yields better results than the single objective one
employed in previous work.  On the other, we aim to verify if the
conclusions reached in \citep{EVITAchapter2008} with regard to the
best algorithm to use within EVITA still hold after the improvements
carried out in the TS and ACO algorithms.

For this purpose, we have tested the selected VRP algorithms
described above on each one of the ten instances selected, each with a different geographic layout. We performed 10 runs per VRP algorithm and instance
with a termination criterion in all cases of 100 generations. The motivation for such a small number of runs is the high computational expense of some of the instance-algorithm combinations: the running times ranged from several minutes to several days\footnote{The computers employed were PCs with
 Intel Celeron processor, between 1 and 3GHz, between 256 and 512 MB RAM.} depending on the algorithm and the size of the instance.

The results were evaluated on two fronts: quantitatively for the
total costs obtained, and qualitatively for the computational time
taken in the runs. The latter is important when considering a
possible commercial application of the EVITA methodology.

\section{Results and analysis}\label{sec:results}

\begin{figure}[htbp!]
\begin{center}
%\fbox{
\resizebox{6.5cm}{!}{
\includegraphics{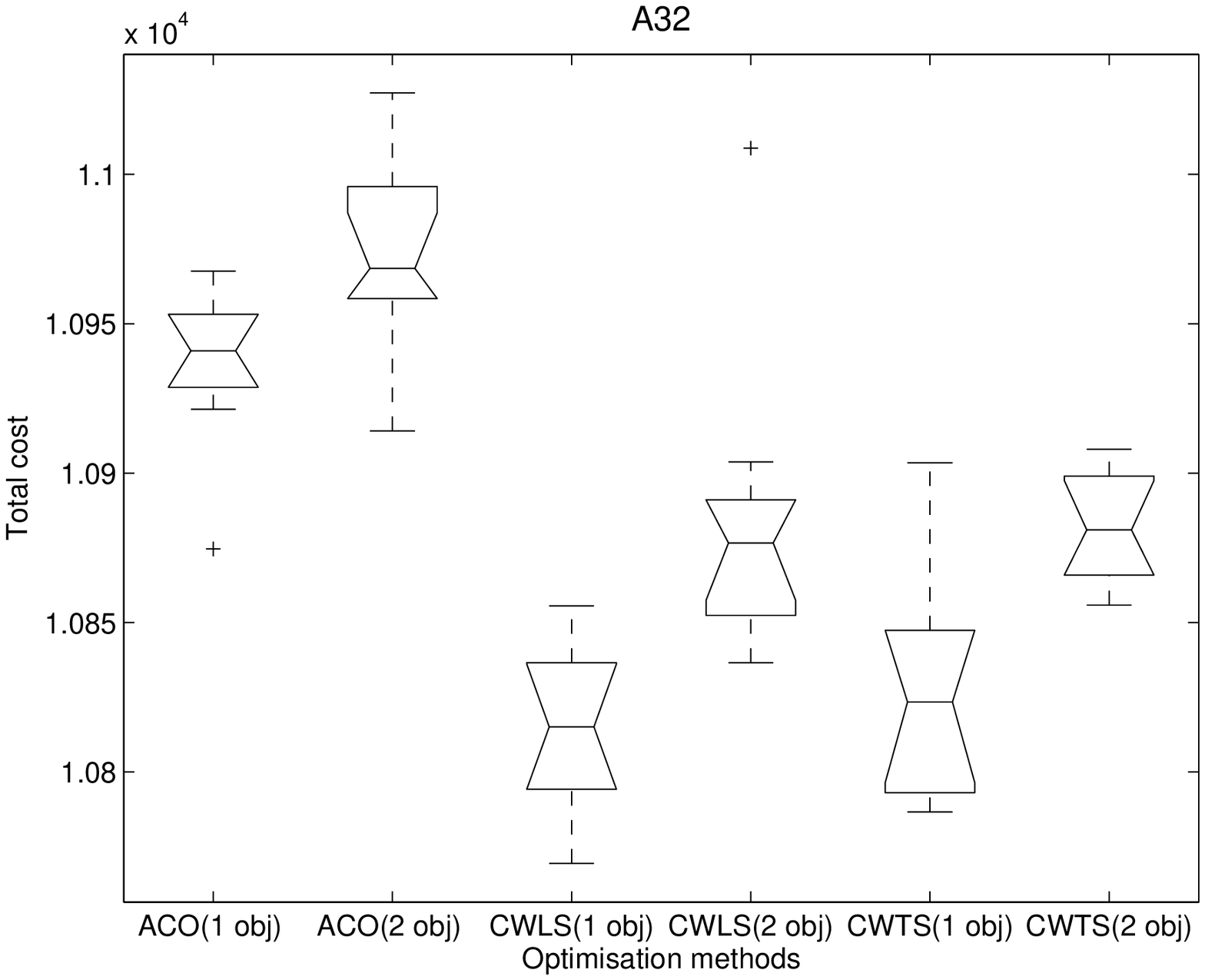}}
\resizebox{6.5cm}{!}{\includegraphics{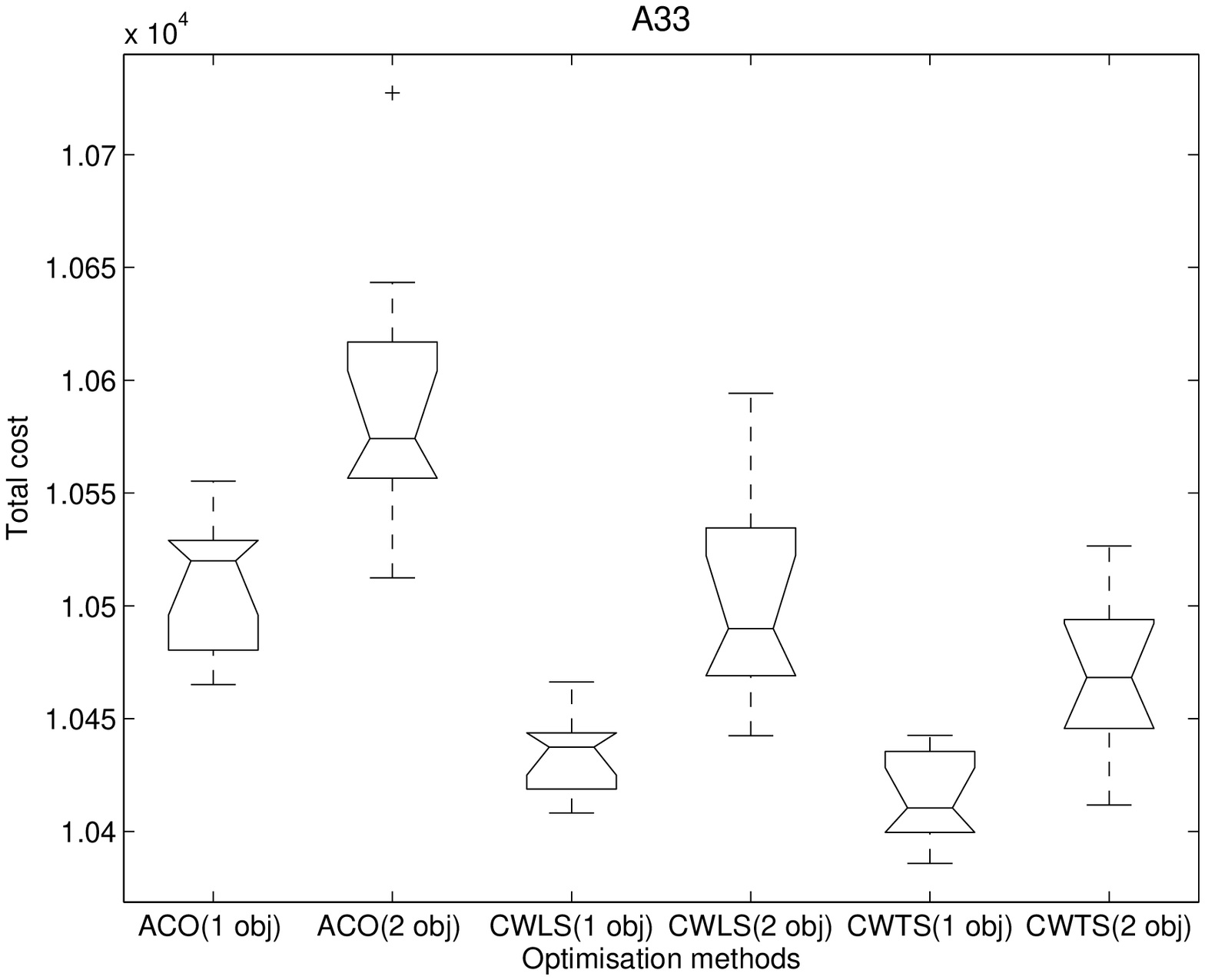} }
\resizebox{6.5cm}{!}{
\includegraphics{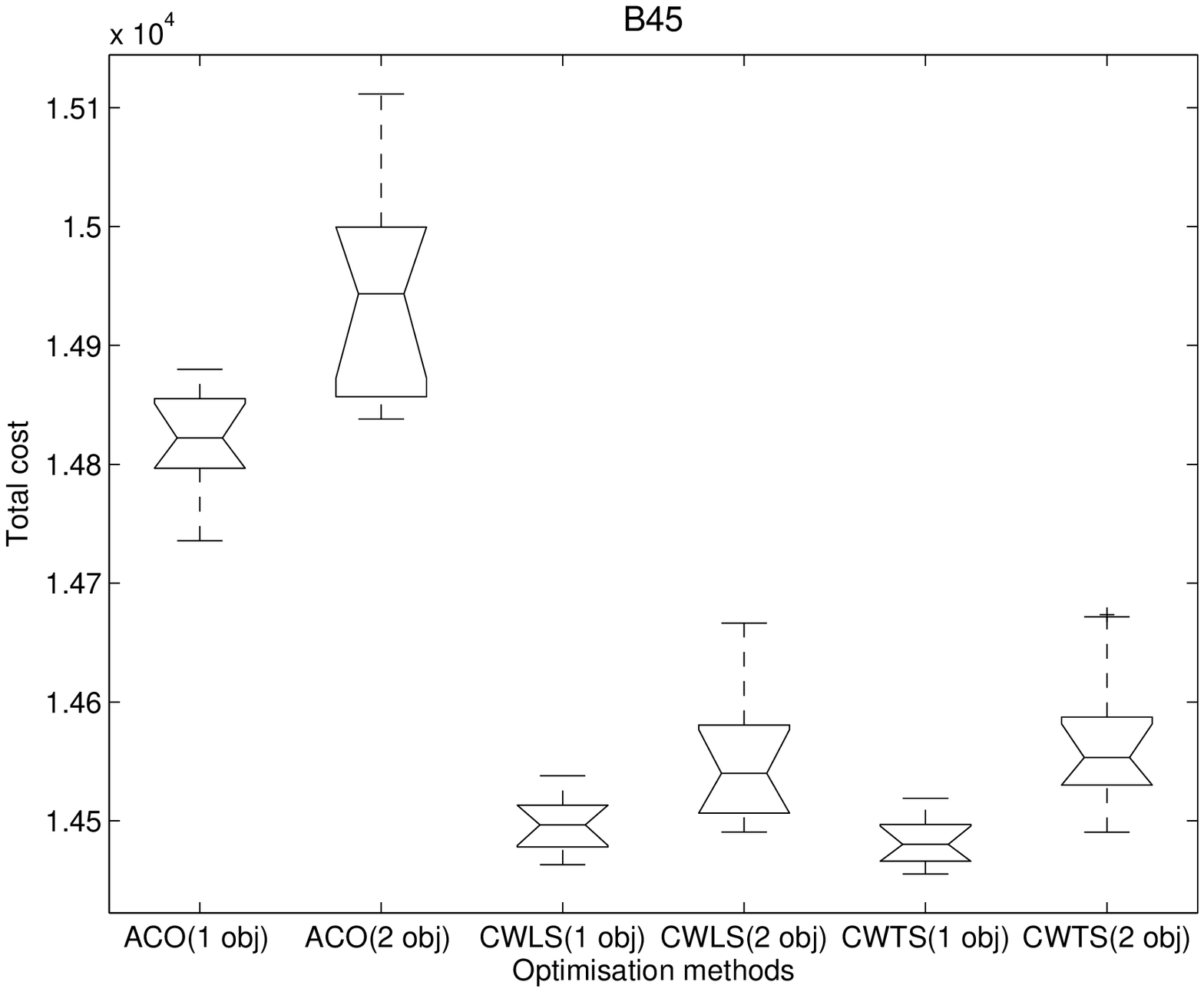}
} \resizebox{6.5cm}{!}{
\includegraphics{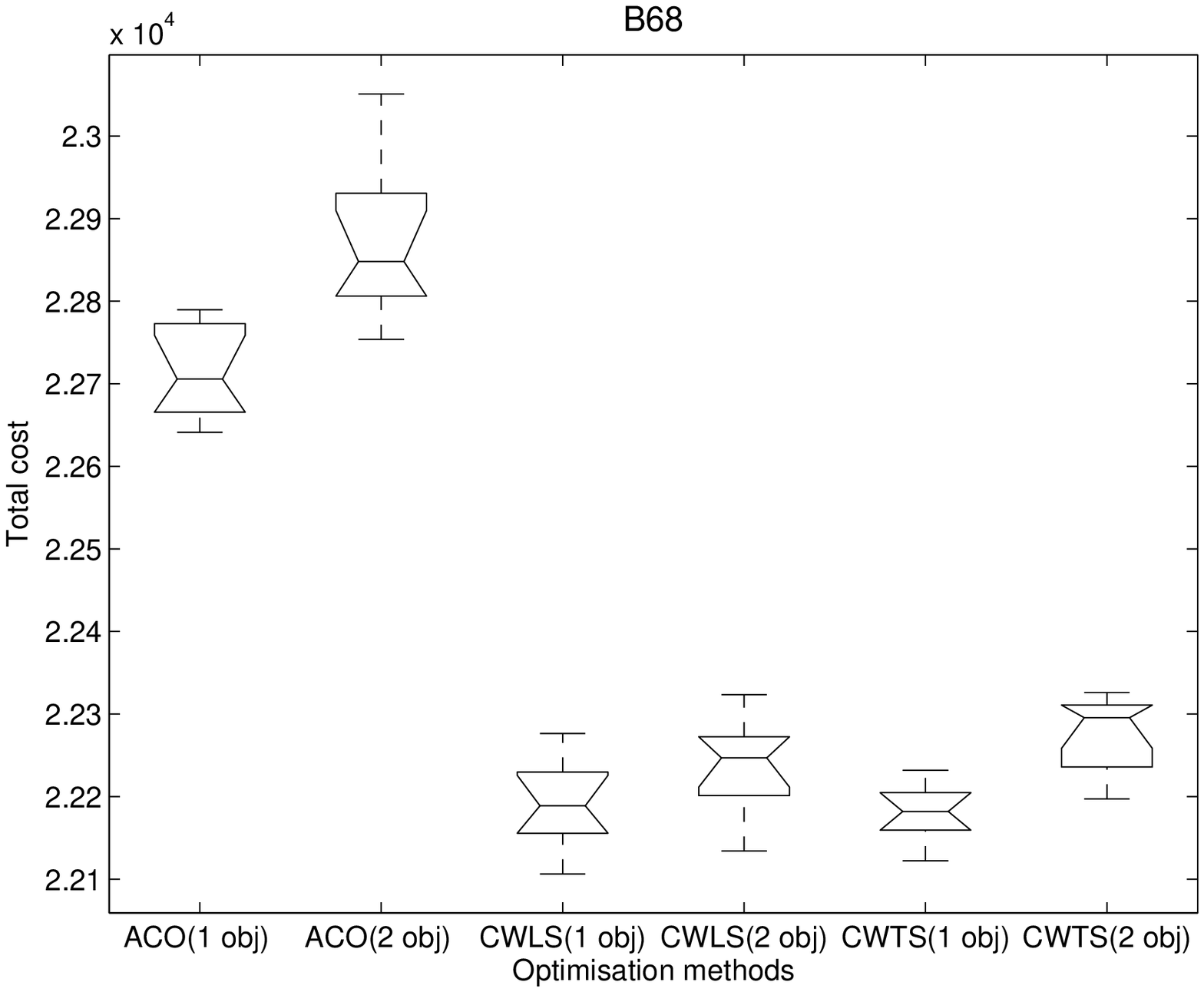}}
\caption{\label{fig:boxplot-smallprobs}\emph{Boxplots} of total
costs for instances A32, A33, B45 and B68.}
\end{center}
\end{figure}

\begin{figure}[htbp!]
\begin{center}
\resizebox{6.5cm}{!}{
\includegraphics{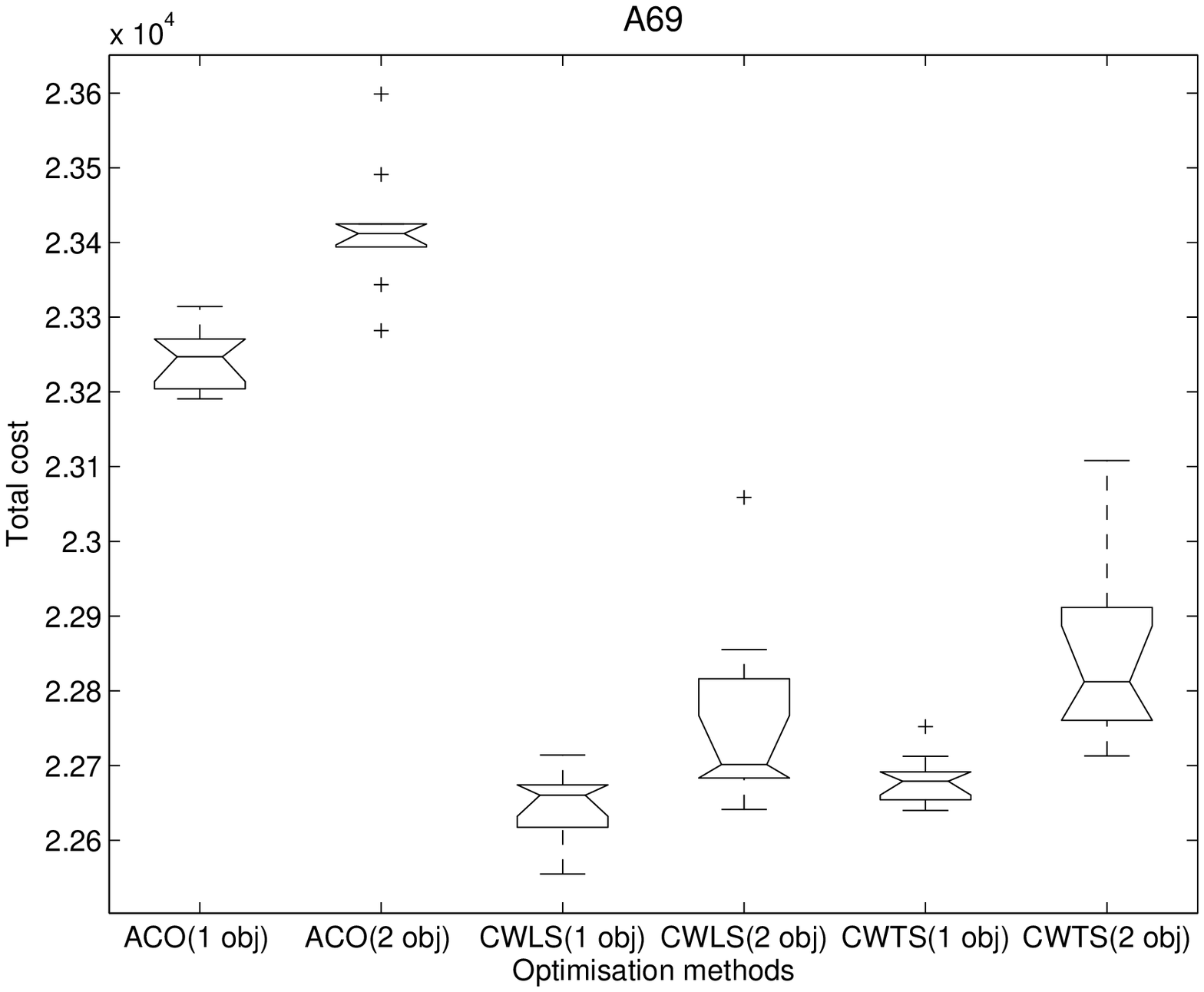}}
\resizebox{6.5cm}{!}{\includegraphics{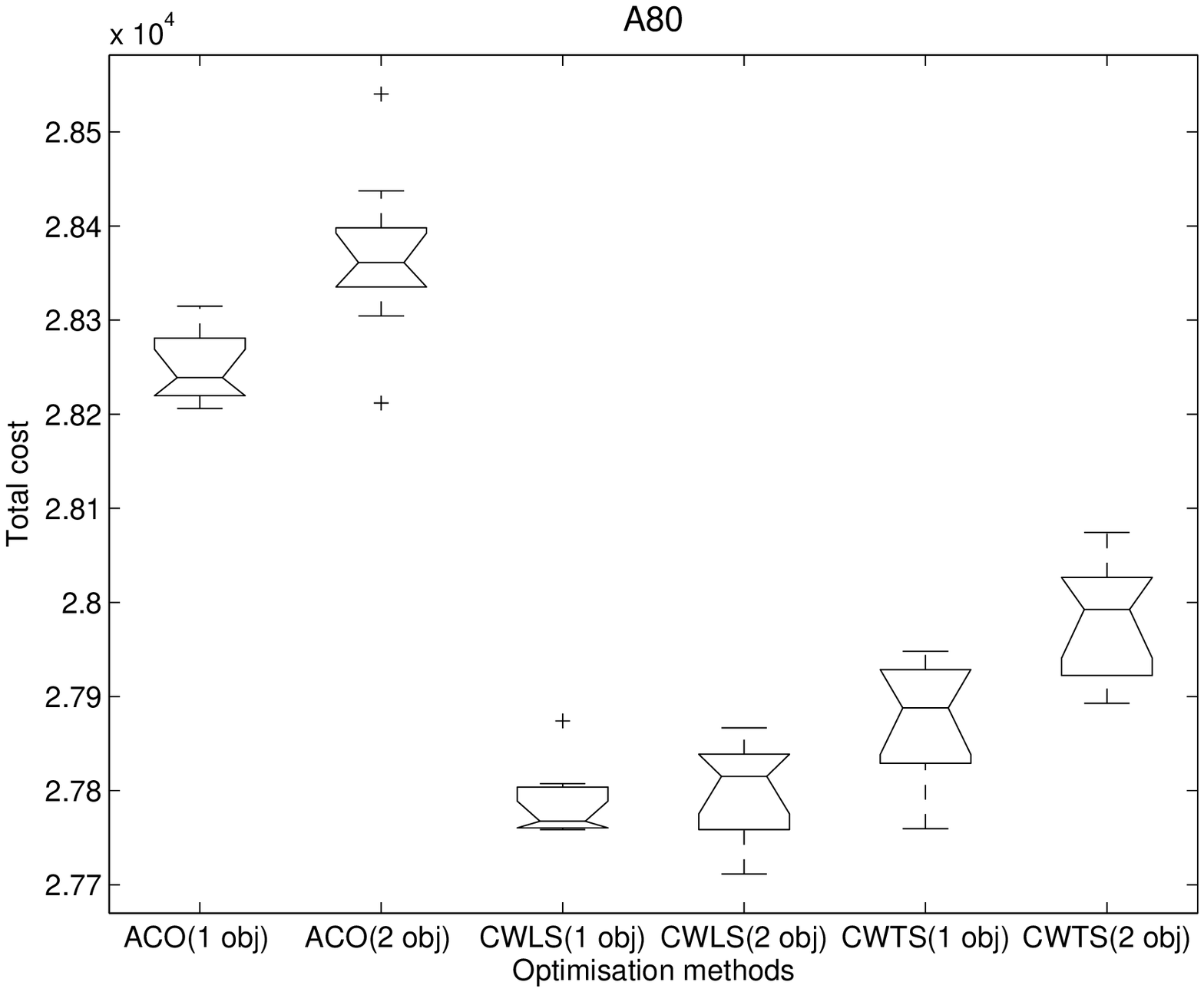}}
\resizebox{6.5cm}{!}{
\includegraphics{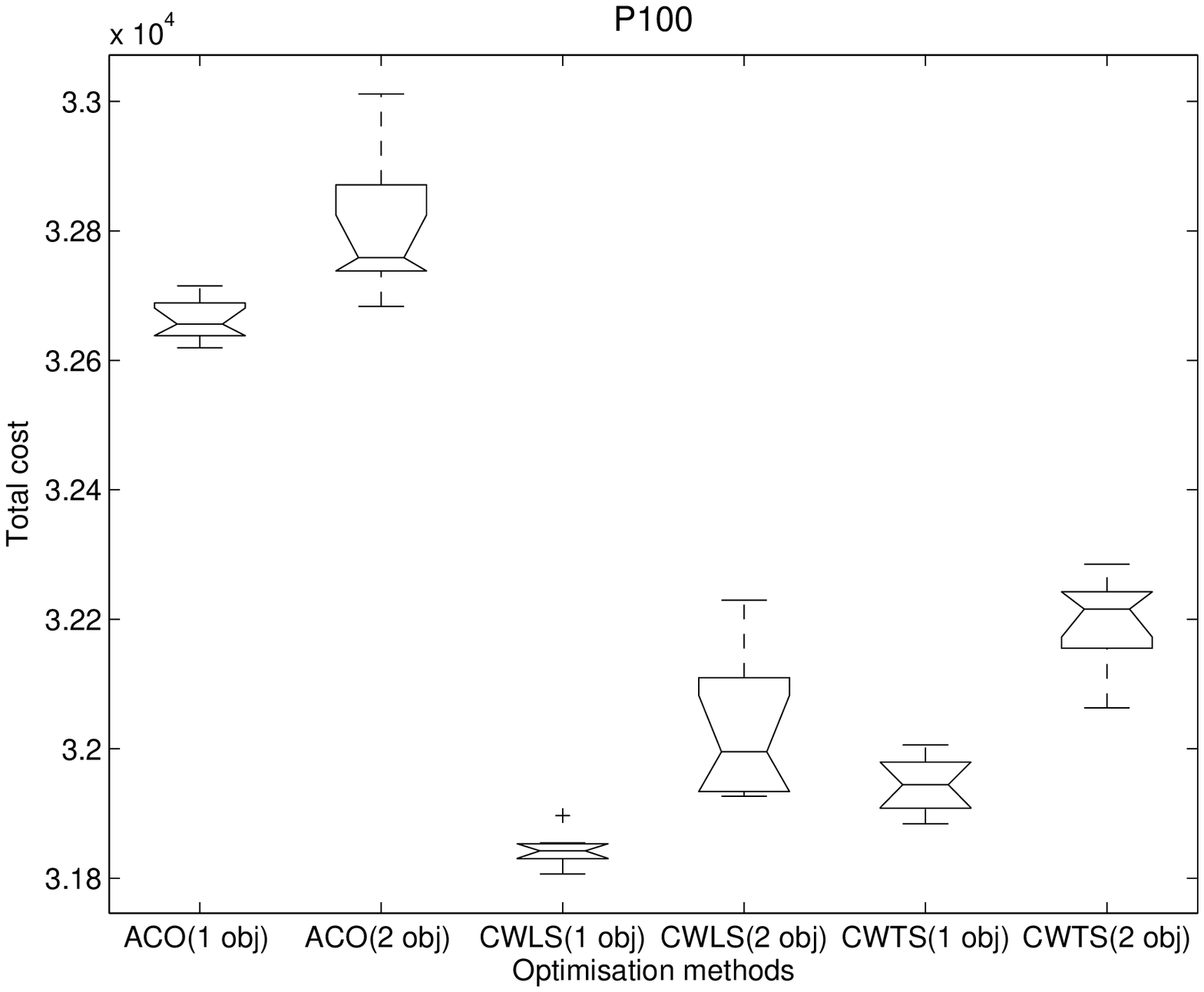}
} \resizebox{6.5cm}{!}{
\includegraphics{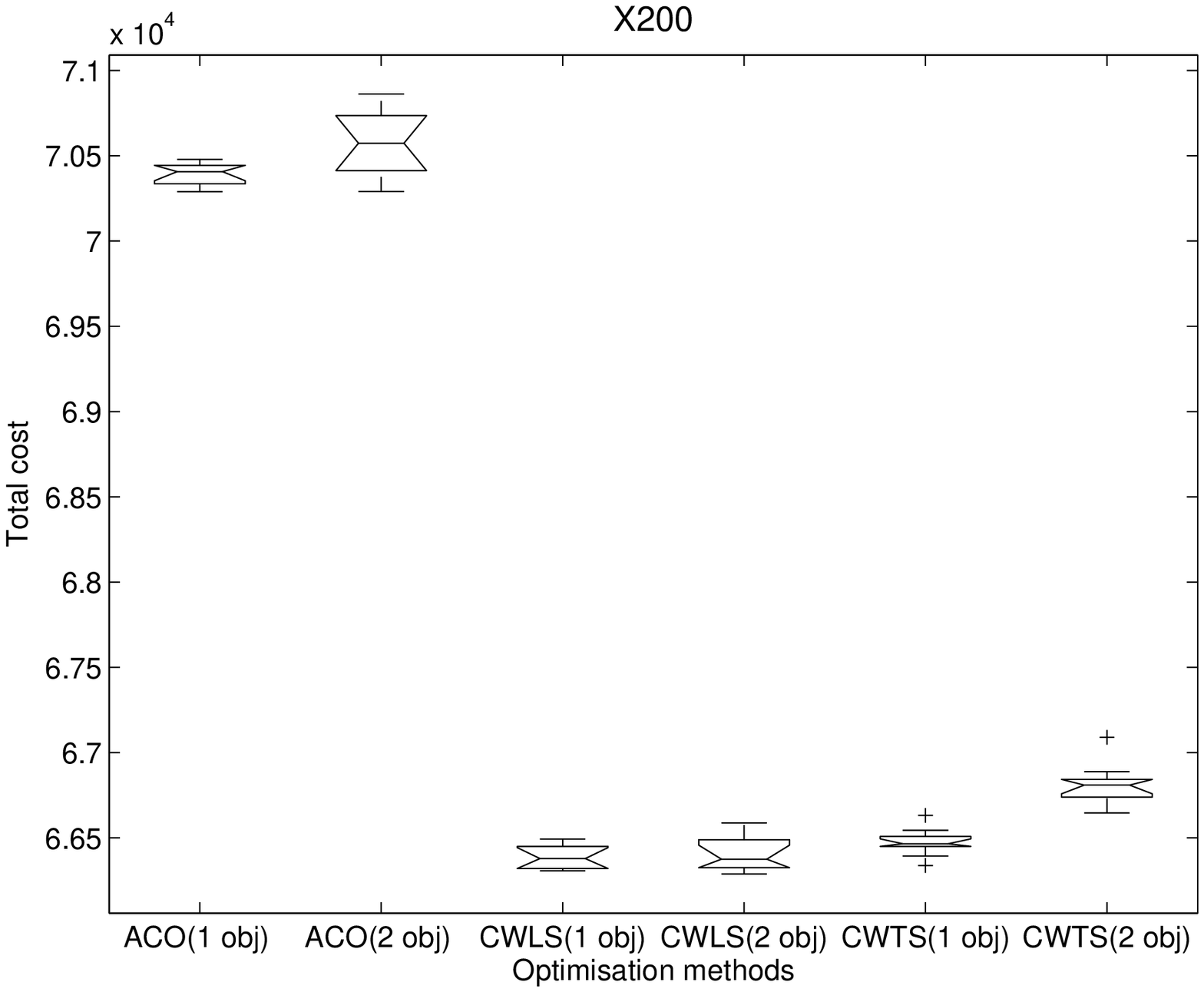}
} \caption{\label{fig:boxplot-bigprobs}\emph{Boxplots} of total
costs for instances A69, A80, P100 and X200.}
\end{center}
\end{figure}

\begin{figure}[htbp!]
\begin{center}
%\resizebox{7cm}{!}{
%\includegraphics{figures/B67.eps}}
\resizebox{7cm}{!}{
\includegraphics{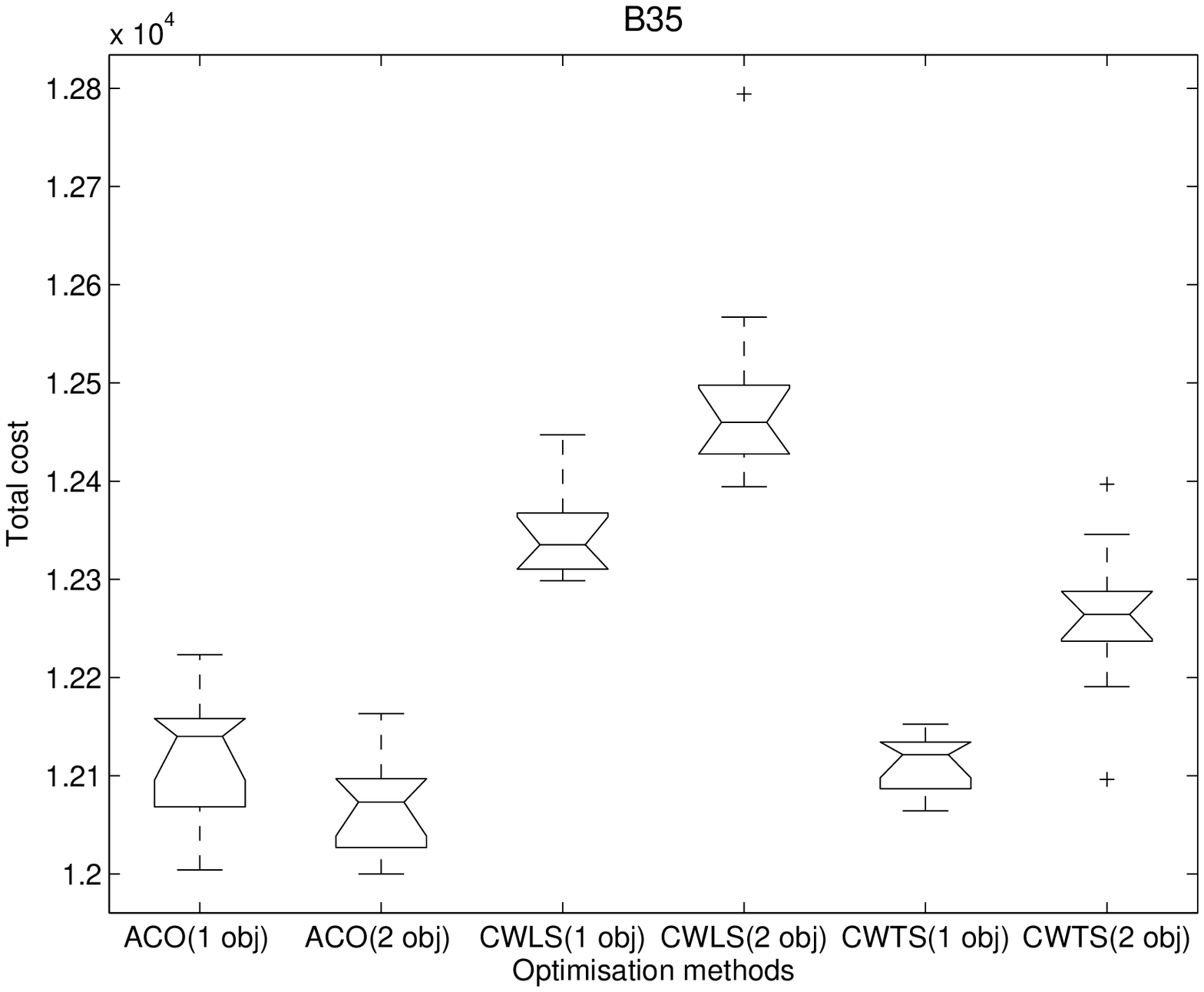}} \\
%\resizebox{7cm}{!}{\includegraphics{figures/A69.eps}} \\
\resizebox{7cm}{!}{
\includegraphics{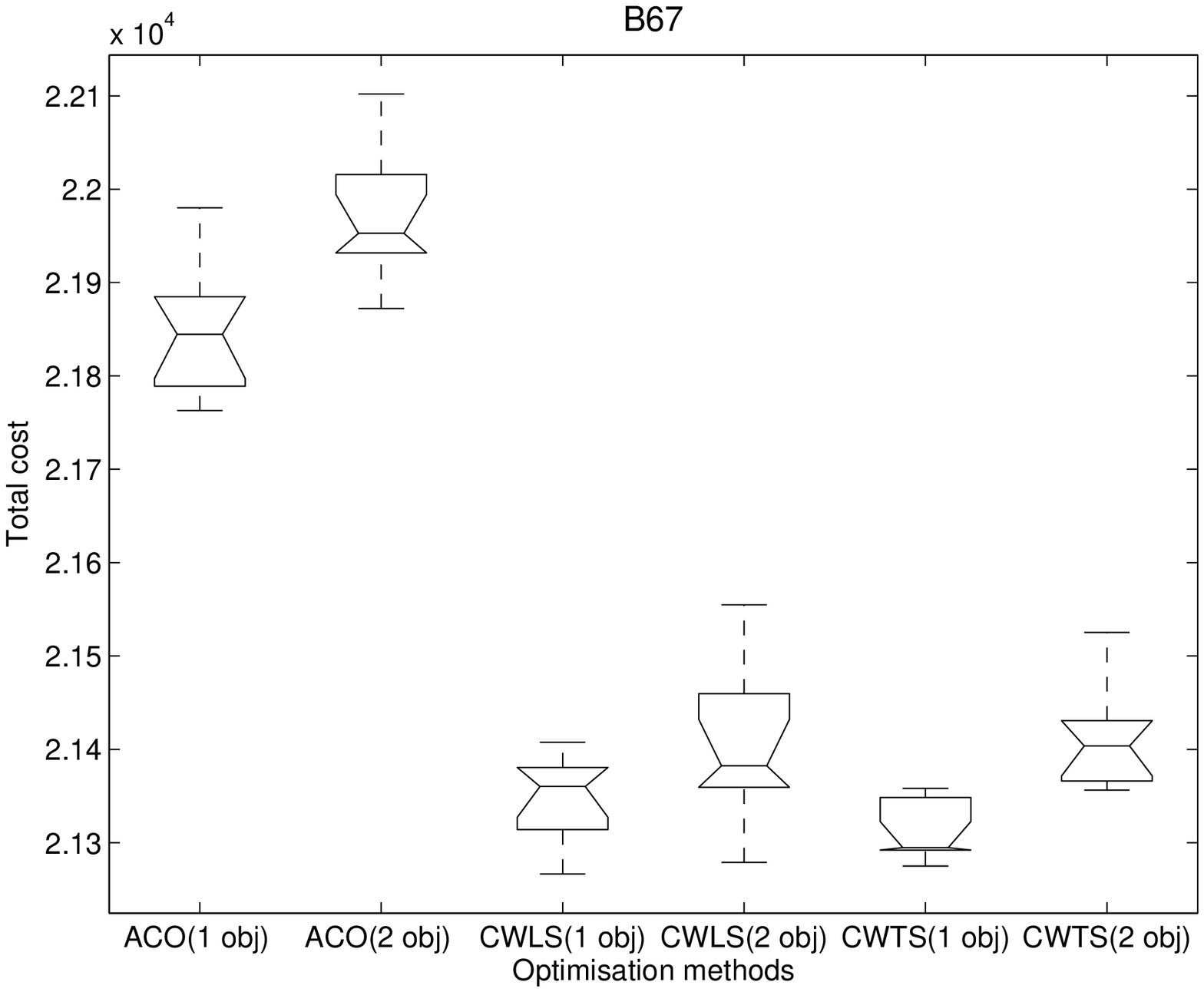}
 }
\caption{\label{fig:boxplot-rareprobs}\emph{Boxplots} of total costs
for instances B35 and B67}
\end{center}
\end{figure}

We carried out Kruskal-Wallis tests for the \textbf{total costs} yielded by the best individuals for
all runs and problem instances, both in single and multiobjective. In the multiobjective case, we define the best individual as that member of the final Pareto front yielding the lowest total cost (as defined for the single objective problem)
The Kruskal-Wallis test is a non-parametric test for multiple comparisons which is suited to the case at hand, in which the number of runs performed for each combination of instance and VRP solver is small. This test does not require normality or homoskedasticity, which are not
guaranteed in our case. Figures \ref{fig:boxplot-smallprobs},
\ref{fig:boxplot-bigprobs} and \ref{fig:boxplot-rareprobs} show the resulting boxplots for the ten
instances studied.

The boxplot consists of a box and whisker plot for each algorithm.
The box has lines at the lower quartile, median, and upper quartile
values. The whiskers are lines extending from each end of the box to
show the extent of the rest of the data. Outliers are data with
values beyond one standard deviation.

The conclusions that can be reached after analysis of the tests are
as follows:
\begin{itemize}
\item{The single objective approach yields the best performance in all instances.}
\item{Considering separately the multi and single objective runs, in general there are no significant differences
between CWLS and CWTS, although at first sight it would seem that
CWLS performs better than CWTS on instances A32, A69, A80, P100 and
X200 and vice versa on A33, B35, B45 and B67. On instance B68 there
are no differences at first sight.}
\item{ACO is significantly worse in all cases except B35.
Furthermore, it is the method that scales worse, getting worse
results as the size of the problem increases. }
\item{The case of instance  B35 is unique in the sense that monoobjective CWTS
is significantly better than CWLS and does not differ significantly
from ACO, both for single and multiobjective. }
\end{itemize}

In order to be able to compare results between the different instances we normalised the fitness values by defining
the \emph{relative percentage deviation}, $RPD$, given by the
following expression:

\begin{equation}\label{eqn:RPD}
RPD = \frac{fitness - fitness_{min}}{fitness_{min}} \times 100
\end{equation}

where $fitness$ is the fitness value obtained by an algorithm
configuration on a given instance. The $RPD$ is, therefore, the
average percentage increase over the lower bound for each instance,
$fitness_{min}$. In our case, the lower bound is the best result
obtained for that instance across all algorithm configurations.

\begin{figure}[btp!]
\begin{center}
\resizebox{7cm}{!}{
\includegraphics{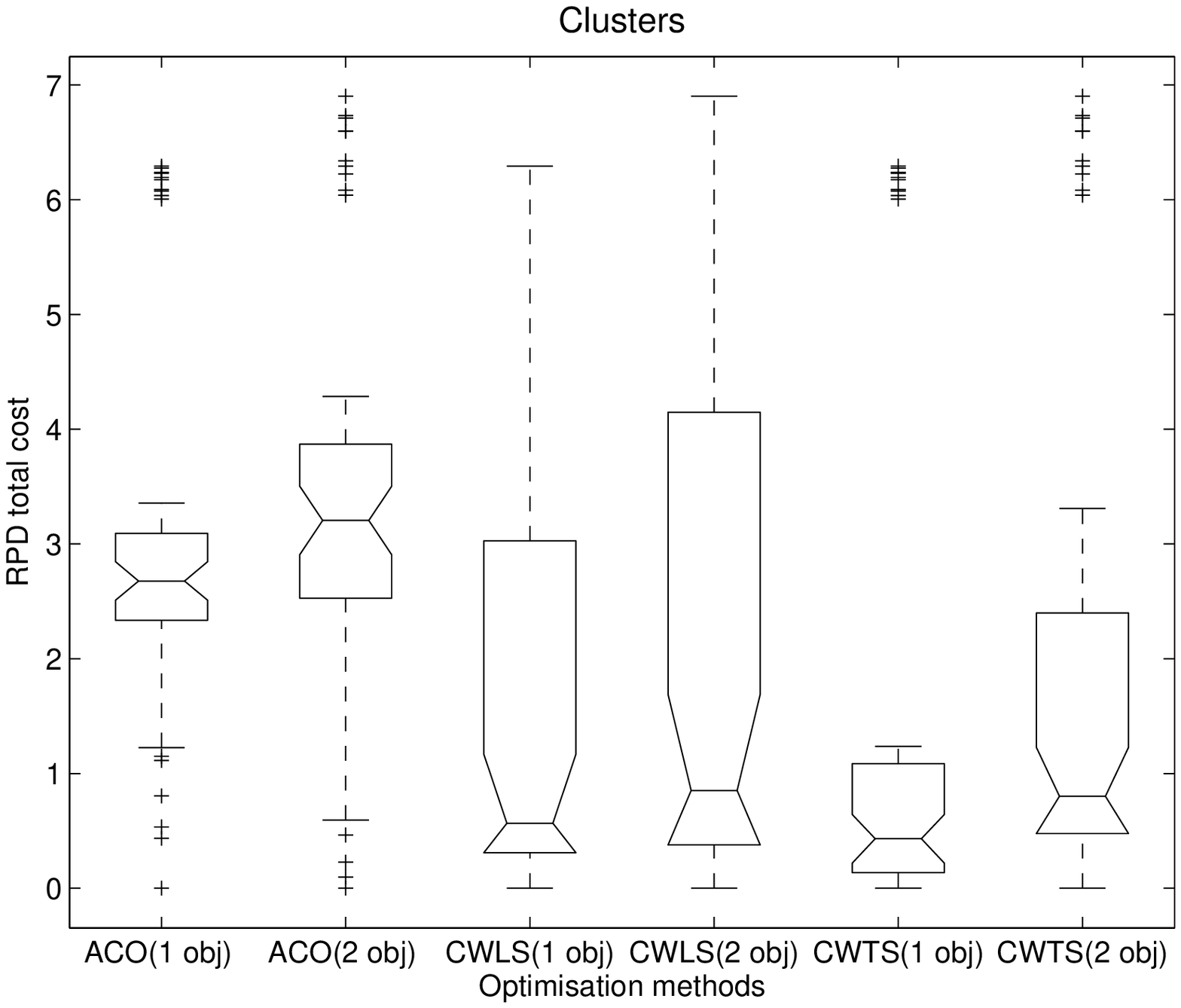}
} \resizebox{7cm}{!}{
\includegraphics{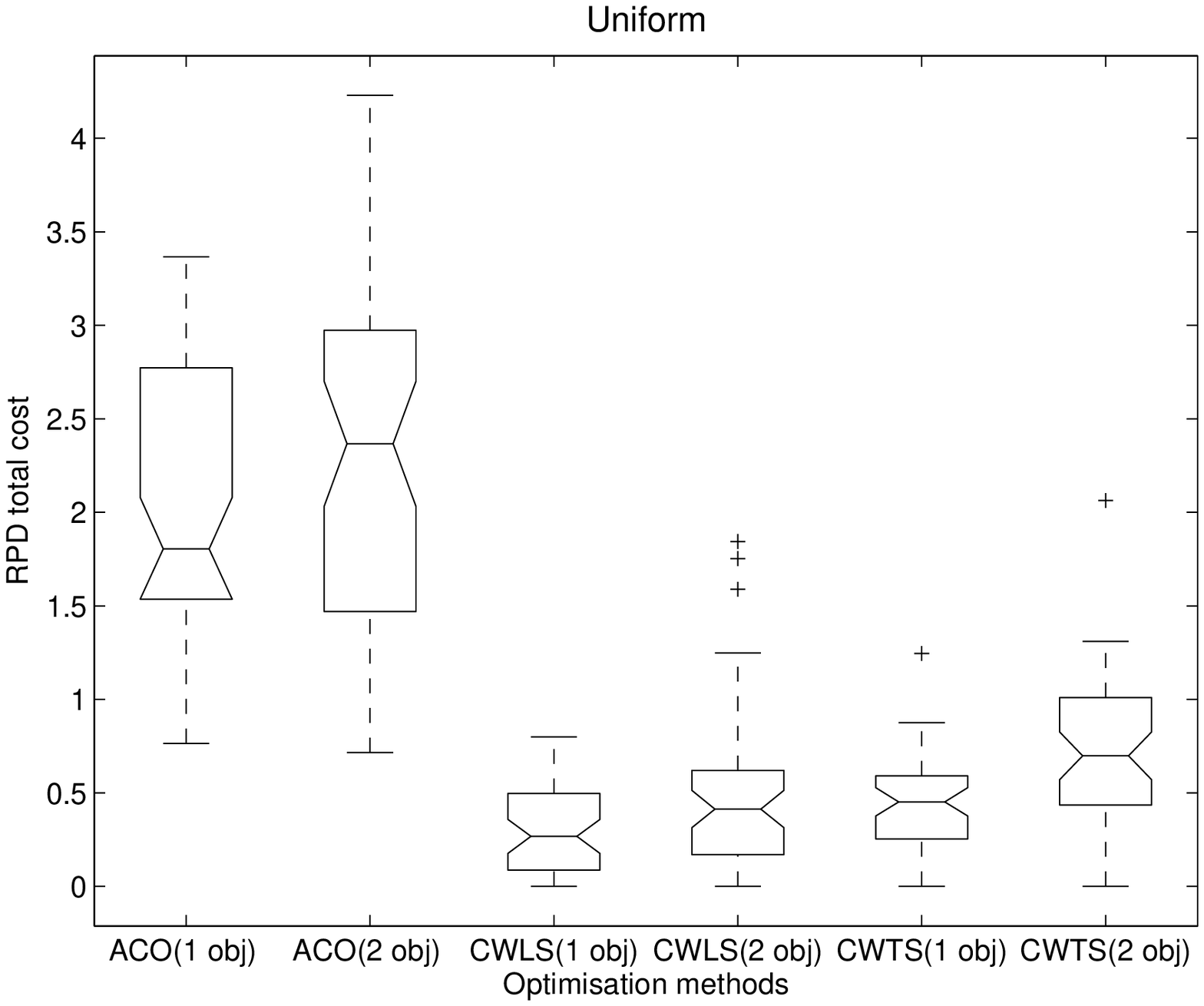}
}\caption{Boxplots of RPD of total costs for groups \emph{clusters}
(top) and \emph{uniform} (bottom)} \label{fig:boxplot-RPD_AB}
\end{center}
\end{figure}

With the $RPD$ results of all the runs for all \emph{VRP solvers} we
ran the tests again; the results are shown in Figure
\ref{fig:boxplot-RPD_AB}  split into two groups: \emph{uniform}
distribution of shops and distribution in \emph{clusters}. The
conclusions in this case are similar. In both groups CWLS and CWTS
perform better than ACO and, at first sight, CWLS is better than
CWTS for group \emph{uniform} and vice versa for group
\emph{clusters}. Further, considering each \emph{VRP solver }
separately, the single objective approach is better than the
multiobjective one.

\begin{figure}[htbp!]
\begin{center}
\resizebox{7cm}{!}{
\includegraphics{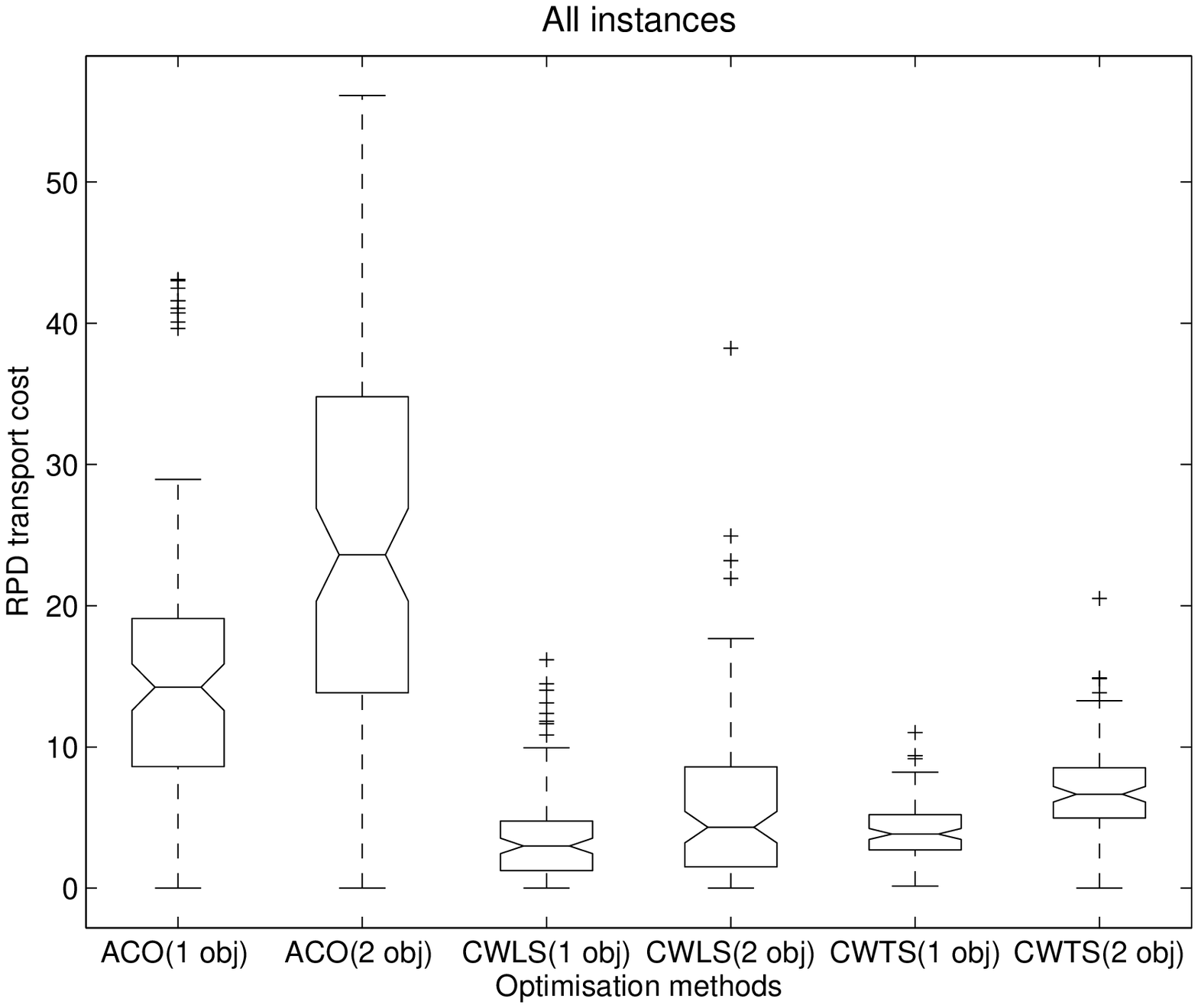}
} \resizebox{7cm}{!}{
\includegraphics{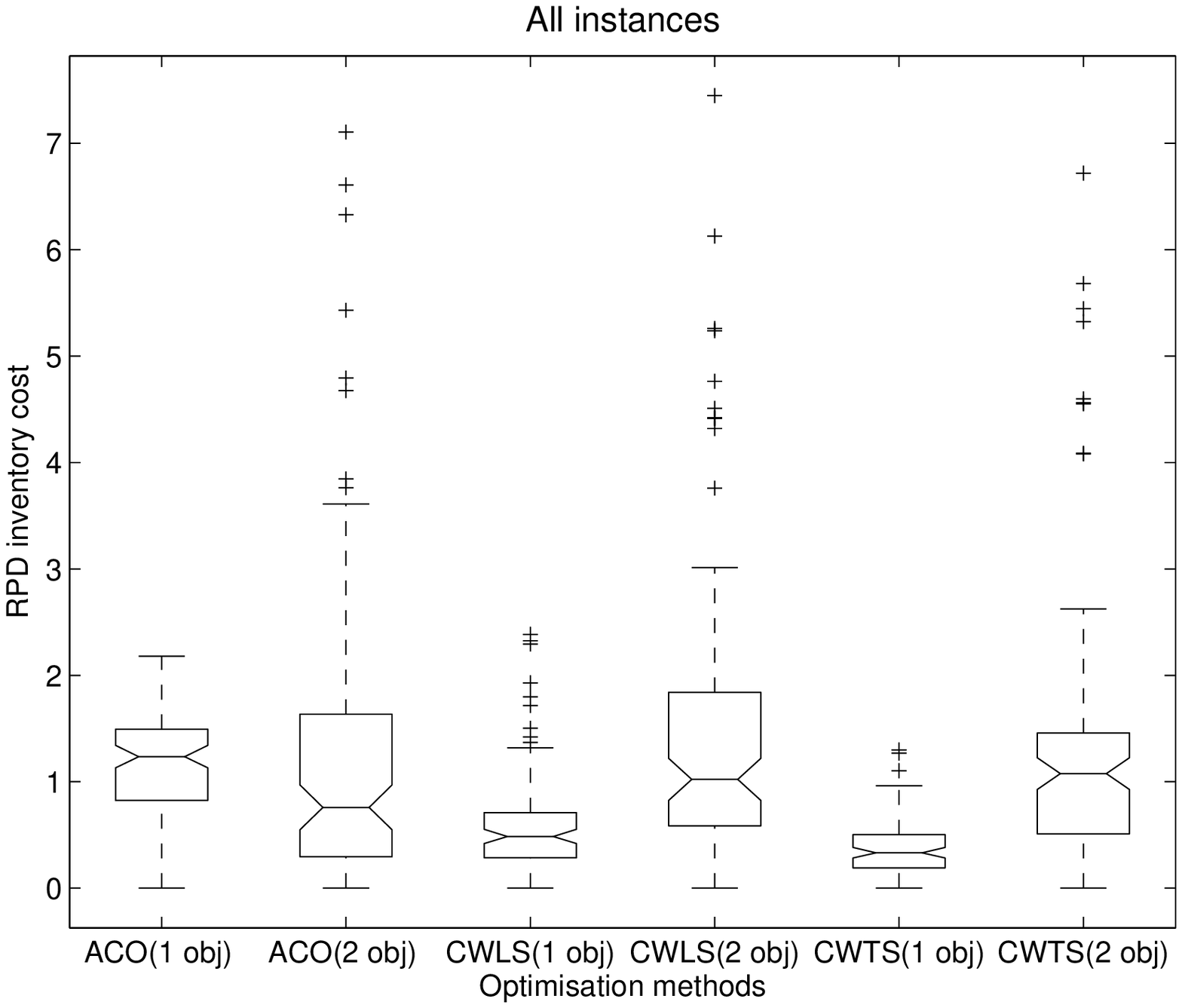}
} \caption{\label{fig:inv_vs_transp}Boxplots for the transport (top)
and inventory costs (bottom)}
\end{center}
\end{figure}

Figure \ref{fig:inv_vs_transp} portrays the comparison between
methods when the costs of transport and inventory are considered
separately. Here it can be observed that there are no significant
differences between methods if only inventory costs are taken into
account. This differs from the conclusions obtained in
\citep{esllucarshaand2006b}, where the choice of VRPsolver
influenced the inventory cost results. In that work, however, the
algorithms employed were suboptimal compared to the ones used here.
So, it could be concluded that the choice of VRPsolver does not have
an influence on the inventory costs \emph{provided a ``good enough"
algorithm is chosen}.

The big difference lies in the transport costs, which is where ACO
clearly shows its inferiority, especially in the multiobjective
approach.

Regarding the computational time, the results clearly favour CWLS
over all other VRP solvers. In general, when employing CWLS the time
for a whole run took approximately the same as that of a single
generation in when using CWTS or ACO. This is a point in favour of
CWLS when considering a potential commercial application.

\subsection{Pareto fronts}\label{subsec:paretoFront}

\begin{figure*}[bt]
\begin{center}
%\fbox{
\resizebox{6cm}{!}{\includegraphics{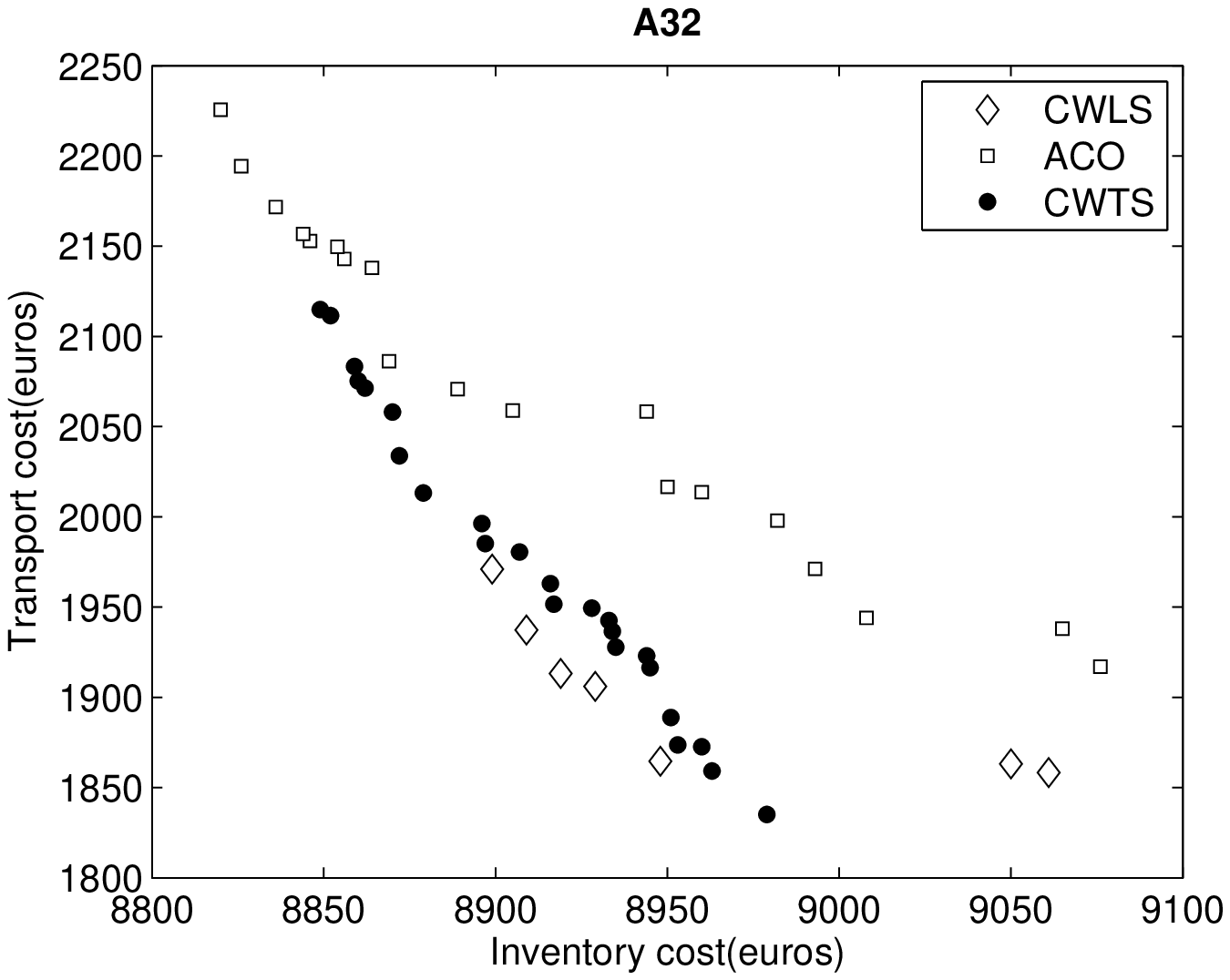}}
\hspace{1.5cm}
\resizebox{6cm}{!}{\includegraphics{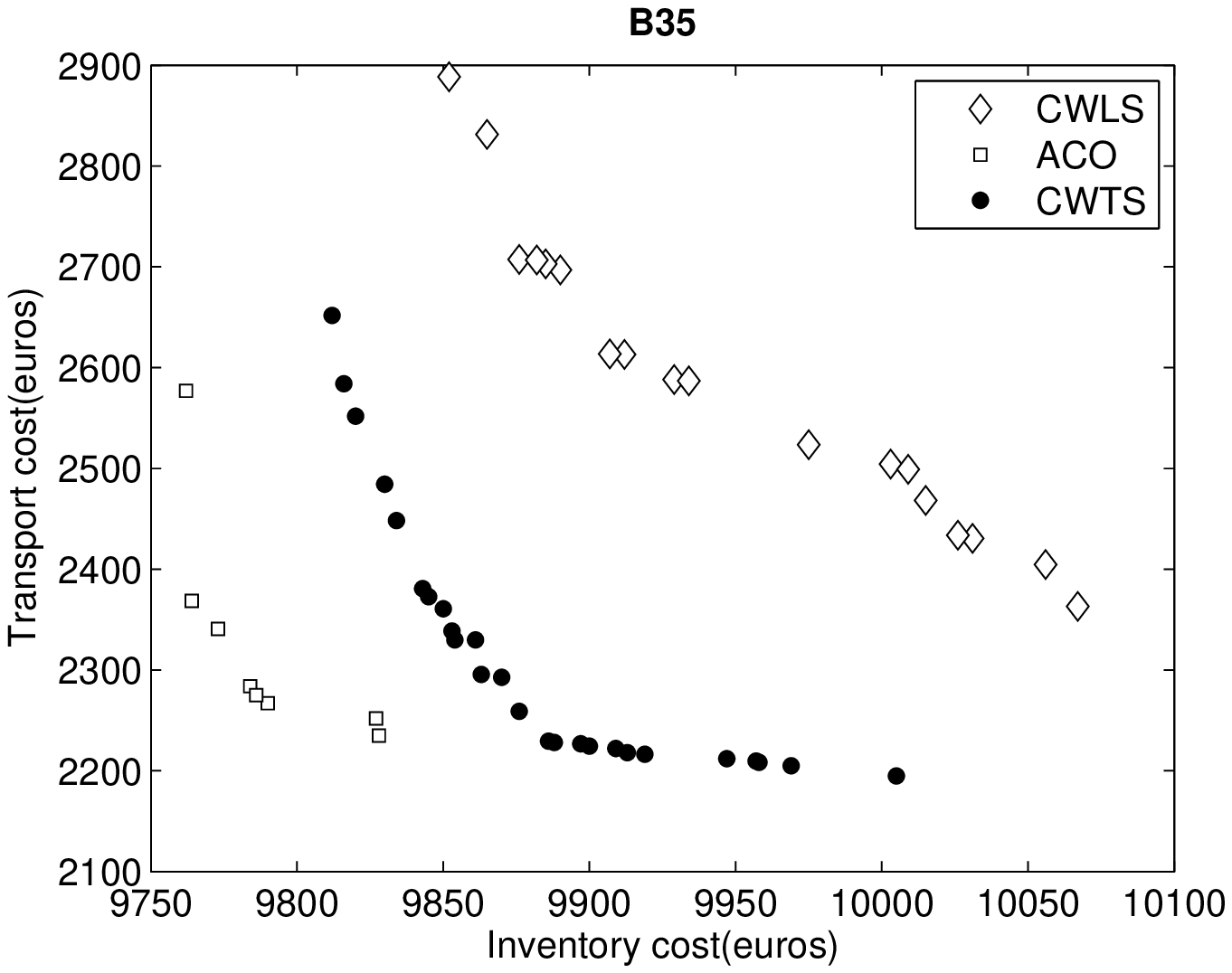}}
 %}
\caption{Pareto fronts in multi-objective approach for one run of
instances A32 (left) and B35 (right) using CWLS, ACO and CWTS as
VRP-solvers.} \label{fig:paretoFronts}
\end{center}
\end{figure*}

In this section we study the Pareto fronts obtained in the
multiobjective approach in two instances of the problem, namely B35
and A32. The former has been chosen because of its uncharacteristic
behaviour (as we have seen, in this instance ACO performs better
than the other VRP solvers), and the latter because it is of a
similar size. Pareto fronts for both instances are shown in Figure
\ref{fig:paretoFronts}. In Table \ref{tab:population} we show the
number of non-dominated solutions vs. the number of different
individuals for each algorithm.

\begin{table*}[htb]
  \centering  \caption{Final population characteristics in multi-objective approach for one run of instances
  A32 and B35 using CWLS, ACO and CWTS as VRP-solvers. Population size is 100 individuals.}
  \label{tab:population}
  \begin{tabular}{|l|c|c|c|} \multicolumn{4}{c}{}\\
    \hline \multicolumn{4}{|c|}{Instance A32}\\
    \hline
   % \hline & \parbox[c][1.5cm][c]{2.5cm}{No. of different\\ individuals} & Pareto front size& Ratio \\
        \hline & No. of different individuals & Pareto front size& Ratio \\
    \hline CWLS & 7 & 7 & 1 \\
     ACO & 100 & 19 & 0.19 \\
     CWTS & 24 & 24 & 1 \\
    \hline
  \end{tabular}\\
  \begin{tabular}{|l|c|c|c|}
    \hline \multicolumn{4}{|c|}{Instance B35}\\
    \hline
    \hline & No. of different individuals & Pareto front size & Ratio \\
    \hline CWLS & 18 & 18 & 1 \\
     ACO & 100 & 8 & 0.08 \\
     CWTS & 36 & 26 & 0.72 \\
    \hline
    \multicolumn{4}{c}{}\\
  \end{tabular}

\end{table*}

Comparing the number of non-dominated solutions in the final
generation for each instance, we observe that the Pareto front
yielded by the best performing VRP solver (CWLS in A32 and ACO in
B35) has always a smaller size, whilst the one corresponding to the
worst performing VRP solver has a bigger size (ACO in A32 and CWLS
in B35). The figures also  hint at the possibility that using CWLS
or ACO as the VRP solver causes the Pareto front to converge to a
very reduced set of solutions, while when using CWTS more
non-dominated solutions are preserved.

On the other hand the ratio of repeated individuals over the
population size is very high when using CWTS or CWLS, whilst in ACO
there are no repeated individuals. It is possible that algorithm ACO
has a slower rate of convergence than the other two in most of
instances (except in cases such as B35), so it requires more
generations to obtain equivalent results.

At this point it would be of interest to measure the quality of the
Pareto front using one of the metrics that have been proposed for
this purpose \citep{coello05}. However, many of them (e.g. the
\textit{error ratio} or the \textit{generational distance}) assume
that knowledge exists on the actual Pareto front, which is not the
case here. Other metrics measure the distribution of solutions on
the Pareto front by evaluating the variance of neighboring
solutions. An example of this is the \textit{spacing}, $SP$, which
measures the relative distances between the members of Pareto front;
a value of $SP = 0$ means that members of the Pareto front are
equispaced. The spacing is given by the following equation:

\[ SP = \sqrt{ \frac{1}{n-1} \sum_{i=1}^n \left( \overline{d} -
d_i\right)^2 } \]

where $n$ is the number of non-dominated solutions found, the
distance $d_i$ is given by

\[d_i = min_j (|f_1^i(x)
-f_1^j(x)|+|f_2^i(x) -f_2^j(x)|), \qquad i,j=1,...,n\]

where
$f_N^k(\cdot)$ is the fitness of point $k$ on objective $N$, $x$ is
the generation number and $\overline{d}$ is the mean of all $d_i$.

The values obtained are given in Table
\ref{tab:pareto_spacing_values}. From these we can see that the best
values of the metric (i.e. the lowest spacing) are obtained by CWTS;
however, from previous analysis we know that it is the other two VRP
solvers that perform better: ACO for B35 and CWLS for A32. We can
hence conclude that this metric is not very meaningful for the
purposes of our problem.

\begin{table}[htb]
\begin{center}\setlength{\tabcolsep}{3mm}\caption{\label{tab:pareto_spacing_values}Values of the spacing for the Pareto fronts generated by all VRP solvers}
{\footnotesize
\begin{tabular} {|l|r|r|} \multicolumn{3}{c}{}\\ \hline
&\multicolumn{2}{c|}{ }\\
 & \multicolumn{2}{c|}{Problem instance} \\
 &\multicolumn{2}{c|}{ }\\
 \cline{2-3}
\emph{VRP solver} & \textbf{A32} & \textbf{B35} \\ \hline & & \\
CWLS  & 17.605 & 23.937\\
ACO & 12.823 & 68.027\\
CWTS  & 8.394 & 17.858\\
& & \\ \hline
 \multicolumn{3}{c}{ }\\
\end{tabular}}\end{center}
\end{table}

%spacing(paretoCWLS32) = 17.6046
%spacing(paretoHybridACO32) = 12.8227
%spacing(paretoTS32) = 8.3942
%spacing(paretoCWLS35) = 23.9368
%spacing(paretoHybridACO35) = 68.0270
%spacing(paretoTS35) = 17.8578

\section{Conclusions and future work}\label{sec:conclusions}

We have shown how, for the problem presented here, the
multiobjective approach does not yield any advantage over the single
objective one. This could be explained by the fact that inventory
costs are well above the transport costs. Given that the
multiobjective approach does not prefer one objective over the
other, it can happen that there are solutions for which transport
costs are very low, but still that does not compensate for high
inventory costs.

Further, we have shown how a classical algorithm such as Clarke and
Wright's, enhanced with local search, can be the best choice in the
context of the Inventory and Transportation Problem, both in terms
of the quality of the solutions obtained and the computational time
necessary to achieve them. The power of other algorithms known to
perform well in the context of VRP, such as ACO and TS, does not
grant a good performance for the joint inventory and transportation
problem. In general, using a global optimisation algorithm such as
evolutionary computation jointly with a heuristic method adapted to
the problem at hand, such as CWLS, yields the best results, so this
is no surprise.

It could be argued that both TS and ACO require a finer tuning of
their parameters in order to give an adequate performance than what
was achieved here. This, however, could be interpreted as a
disadvantage of their application to a variety of problem
configurations and in a commercial context.

\begin{figure}[htbp!]
\begin{center}
\resizebox{8cm}{!}{
\includegraphics{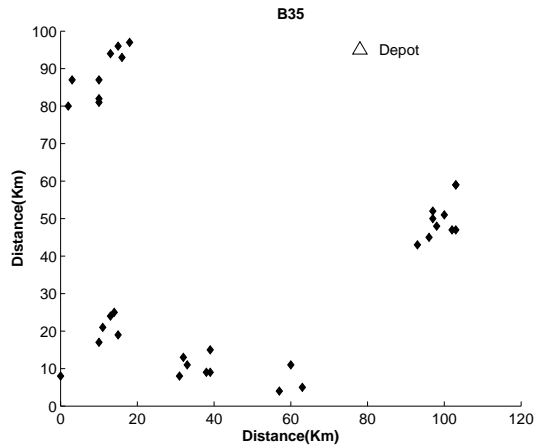}
} \caption{\label{fig:B35layout} Geographical layout for instance
B35}
\end{center}
\end{figure}

Special attention should be given to the case of instance B35, since
it is the only one for which ACO yields better results than the
remaining VRP solvers (with the exception of single objective TS)
and, oddly, this happens in the multiobjective case. The
geographical layout of this instance is shown in Figure
 \ref{fig:B35layout}; the high eccentricity of the distribution can be
 seen (the depot is on one side of the shops) compared to the small
 number of shops. Instance A80 also has a similar value of
 eccentricity, but for a higher number of shops. Perhaps here lies
 the explanation of why ACO works better in the former but not in
 the latter; clarifying this point is left for future work. In any
 case, we can conclude that although the CWLS with single objective
 approach performs better in general, it is nonetheless interesting
 to have a tool that can provide the possibility of using other VRP
 solvers and a multiobjective approach in order to handle special
 cases, such as B35.

It must be noted that this result is subject to the specificities of
the problem data, i.e. the fact that in this case the inventory cost
greatly outweighs the cost of the transport. As future work we must
consider a case in which the products moved are cheaper and hence
the inventory cost is more in a par with the transport cost.

%\nocite{*}
%\nocite{barajas}

\bibliographystyle{apalike}
\bibliography{EVITA}

\section*{Acknowledgements}

This work was part of project \emph{NoHNES - Non Hierarchical
Network Evolutionary System}, and has been supported by the Spanish
Ministry of Science and Innovation, ref. TIN2007-68083-C02.

\section*{Appendix I: NSGA-II}\label{app:nsga2} NSGA-II \citep{nsgaii02, nsgaII2000} is an non-elitist
multiobjective evolutionary algorithm (MOEA) which was developed
 in order to overcome the problems of previous MOEAs, such as the high computational
complexity of sorting non dominated solutions. These algorithms have
two common features: assigning fitness to population members based
on nondominated sorting and preserving the diversity among solutions
of the same nondominated front.

NSGA-II works as follows: Initially, a random parent population
$P_0$ is created, with size N. The population is sorted based on
nondominance. Each solution is assigned a fitness (or rank) equal to
its nondomination level (with 1 being the best level). Initially,
the usual binary tournament selection, recombination, and mutation
operators are used to create an offspring population $Q_0$ of size
N. For the remaining generations $t$ we do the following: First, a
population $R_t$ of size 2N is formed as the union of $P_t$ and
$Q_t$ and sorted according to nondomination in a number of fronts
$F$. Next, a new population $P_{t+1}$ of size N is created by
selecting individuals from $R_t$ in order of nondominance (i.e.
ordered individuals from $F_1$ are chosen first, then ordered
individuals from $F_2$ and so on until the number of individuals
belonging to $P_{t+1}$ is N).

The advantages of NSGA-II with respect to previous MOEAs are the
fast sorting of nondominated individuals and the preservation of
diversity.

\paragraph*{Nondominated sorting.} First, for each solution $p$ we
calculate two entities: domination count, $n_p$, i.e. the number of
solutions which dominate $p$, and a list of solutions that $p$
dominates, $S_p$. All solutions in the first nondominated front will
have their $n_p$ as zero. Now, for each solution with $n_p=0$, we
visit each member $q$ in its $S_p$ and reduce its domination count
by one. In doing so, if for any member the domination count becomes
zero, we place it in a separate list $Q$. These members belong to
the second nondominated front. This procedure is continued with each
member of $Q$ and the third front is identified. This process
continues until all fronts are identified. The code for this
operation can be found in Algorithm \ref{alg:NSGAIIfastSorting}.

\begin{Algorithm}[h!]
\begin{center}\setlength{\tabcolsep}{2mm}
\renewcommand{\arraystretch}{1.2}
{\footnotesize \textsf{
\begin{tabular}{l} \multicolumn{1}{c}{}\\ %\hline
\textbf{Procedure} fastNonDominatedSort() \\
\textbf{input:} population $P$\\
\textbf{output:}  listOfFronts $F$ \\
\textbf{For} each $p \in P$\\
\qquad $S_p = \emptyset$ \\
\qquad $n_p = 0$ \\
\qquad \textbf{For} each $q \in P$ \\
\qquad \qquad \textbf{if} $p$ dominates $q$\\
\qquad \qquad \qquad Adds $q$ to $S_p$\\
\qquad \qquad \textbf{else if} $q$ dominates $p$\\
\qquad \qquad \qquad Increments $n_p$\\
\qquad \textbf{if} $n_p = 0$ \\
\qquad \qquad $p.rank = 1$\\
\qquad \qquad Adds $p$ to $F_1$\\
\qquad $f = 1$\\
\qquad \textbf{while} $F_f \not= \emptyset$ \\
\qquad \qquad $Q = \emptyset$ \\
\qquad \qquad \textbf{for} each $p \in F_f$ \\
\qquad \qquad \qquad \textbf{for} each $q \in S_p$ \\
\qquad \qquad \qquad \qquad Decrements $n_q$ \\
\qquad \qquad \qquad \qquad \textbf{if} $n_q= 0$ \\
\qquad \qquad \qquad \qquad \qquad $q.rank = 1$ \\
\qquad \qquad \qquad \qquad \qquad Adds $q$ to $Q$ \\
\qquad \qquad $F_f = Q$\\
\qquad \qquad Increments $f$\\
\textbf{return} $F$\\
\textbf{end;}  \\
%\hline
\end{tabular}
}}\caption{\label{alg:NSGAIIfastSorting}Nondominated sorting
function for NSGAII.}\end{center}
\end{Algorithm}

\paragraph*{Diversity preservation.} To get an estimate of the density of
solutions surrounding a particular solution in the population, we
calculate the average distance of two points on either side of this
point along each of the objectives. This distance is called the
\textit{crowding distance}, and it is calculated as  shown in
Algorithm \ref{alg:NSGAIIcrowdingDist}. Moreover, a
\textit{crowded-comparison operator},$<_n$, is used in order to
guide the selection process at the various stages of the algorithm
towards a uniformly spread-out Pareto-optimal front.

In the selection process, given two solutions with different
nondomination ranks the one with the lower (better) rank will be
preferred. Otherwise, if both solutions belong to the same front,
then one located in a less crowded region will be preferred.

\begin{Algorithm}[htb!]
\begin{center}\setlength{\tabcolsep}{2mm}
\renewcommand{\arraystretch}{1.2}
{\footnotesize \textsf{
\begin{tabular}{l} \multicolumn{1}{c}{}\\ %\hline
\textbf{Procedure} crowdingDistance() \\
\textbf{input:} population pop[popSize]\\
\textbf{for} $i=1..popSize$ \\
\qquad $pop[i].distance = 0$ \\
\textbf{for} each objective $m$ \\
\qquad Sort $pop$ using $m$\\
\qquad $pop[1].distance = pop[popSize].distance = \infty$\\
\qquad for $i=2..popSize-1$\\
\qquad \qquad $pop[i].distance = pop[i].distance + (pop[i+1].m - pop[i-1].m)/(f_m^{max} - f_m^{min})$\\
\textbf{end;}  \\
%\hline
\end{tabular}
}}\caption{\label{alg:NSGAIIcrowdingDist}Crowding distance
assignment for NSGA-II Algorithm.}\end{center}
\end{Algorithm}

For a complete description of NSGA-II, see \citep{nsgaii02,
nsgaII2000}.

\section*{Appendix II: Algorithms for solving the VRP}\label{app:algorithms4VRP}
Here we describe the algorithms employed as VRPsolvers: C\&W's algorithm, Local Search, ACO and TS, plus other additional functions.

\begin{Algorithm}[tb]
\begin{center}\setlength{\tabcolsep}{2mm}
\renewcommand{\arraystretch}{1.2}
{\footnotesize \textsf{
\begin{tabular}{l}
 \multicolumn{1}{c}{}\\ %\hline
\textbf{Algorithm} CWLS \\
\textbf{input:} shops[nShops], depot \\
\textbf{output:} routes\\
\qquad  [Build $nShops$ initial routes with one shop only] \\
\qquad \qquad \textbf{for} i=1... $nShops$  \\
\qquad \qquad \quad $r_i = (depot, shops[i], depot)$\\
\qquad  [Calculate savings] \\
\qquad \qquad Calculate $s_{i,j}$ for each pair shops[i], shops[j]\\
\qquad \qquad $s_{i,j}= cost_{shops[i]depot} + cost_{depot shops[j]} - cost_{shops[i]shops[j]}$ \\
\qquad  [Best union ] \\
\qquad \qquad\textbf{repeat}\\
\qquad \qquad \quad $s_{i*j*} = \max s_{i,j}$\\
\qquad \qquad \quad Let $r_{i*}$ be the route containing
shops[i]\\
\qquad \qquad \quad Let $r_{j*}$ be the route containing shops[j]\\
\qquad  \qquad \quad\textbf{if} \\
\qquad \qquad \qquad shops[i*] is the last shop in  $r_{i*}$\\
\qquad  \qquad \qquad and shops[j*] is the first shop in $r_{j*}$ \\
\qquad  \qquad \qquad and the combination is feasible \\
\qquad  \qquad \quad\textbf{then} combine $r_{i*}$ and $r_{j*}$ \\
\qquad \qquad \quad delete $s_{i*j*}$; \\
\qquad \qquad\textbf{until} there are no more savings to consider;
\\
%\qquad  [Local search] \\
%\qquad \qquad Improve each route $r_i$  separately\\
%\qquad \qquad Improve considering exchanges between routes\\
\qquad    \textbf{return} \texttt{routes}\\
\textbf{end;}  \\
%\hline
\end{tabular}
}}\caption{\label{CWLS-code}Clarke \& Wright's algorithm (C\&W).}
\end{center}
\end{Algorithm}

\begin{Algorithm}[tb!]
\begin{center}\setlength{\tabcolsep}{2mm}
\renewcommand{\arraystretch}{1.2}
{\footnotesize \textsf{
\begin{tabular}{l}
 \multicolumn{1}{c}{}\\ %\hline
\textbf{Algorithm} LocalSearch\\
\textbf{\textbf{input:}} initialSolution\\
\qquad  [Initialisation] \\
\qquad \qquad \texttt{best} $\leftarrow$  initialSolution \\
\qquad \qquad costBest = costTemp\\
\qquad \qquad [Improving each route]\\
\qquad \qquad \textbf{for} each r in routes(initialSolution) \\
\qquad \qquad \qquad \textbf{for} k=1..Max(sizeOf(r),numberOfNeighbors) \\
\qquad \qquad \qquad \qquad select s1, s2 random shops in r \\
\qquad \qquad \qquad \qquad tempSolution $\leftarrow$ InterchangeShops(initialSolution,s1, s2)\\
\qquad \qquad \qquad \qquad \textbf{if} (calculateCost(tempSolution) $<$ costBest)\\
\qquad \qquad \qquad \qquad \qquad \texttt{best} $\leftarrow$ tempSolution \\
\qquad \qquad \qquad \qquad \qquad costBest = calculateCost(best)\\
\qquad \qquad [Improving pairs of routes]\\
\qquad \qquad \textbf{for} r1 =1..routes(initialSolution) \\
\qquad \qquad \qquad \textbf{for} r2=r1+1..routes(initialSolution) \\
\qquad \qquad \qquad \qquad select s1 random shop in r1 \\
\qquad \qquad \qquad \qquad select s2 random shop in r2 \\
\qquad \qquad \qquad \qquad solTemp $\leftarrow$ InterchangeShops(initialSolution,s1, s2)\\
\qquad \qquad \qquad \qquad \textbf{if} (calculateCost(tempSolution) $<$ costBest)\\
\qquad \qquad \qquad \qquad \qquad \texttt{best} $\leftarrow$ tempSolution \\
\qquad \qquad \qquad \qquad \qquad costBest = calculateCost(best)\\
\qquad \textbf{return} \texttt{best}\\
\textbf{end algorithm;}  \\
%\hline
\end{tabular}
}}\caption{\label{alg:localSearch}Local Search (LS) algorithm.}
\end{center}
\end{Algorithm}

\begin{Algorithm}[tb!]
\begin{center}\setlength{\tabcolsep}{2mm}
\renewcommand{\arraystretch}{1.2}
{\scriptsize \textsf{
\begin{tabular}{l}
 \multicolumn{1}{c}{}\\ %\hline
\textbf{Algorithm} ACO \\
\quad \textbf{input:} shops[nShops], numberOfIterations\\
\quad \textbf{output:} \texttt{routes} \\
\quad [Initialise values] \\
\qquad \texttt{routes} $\leftarrow$ validRandomSolution \\
\qquad globalCost $\leftarrow$ calculateCost(\texttt{routes})\\
\quad [Main loop] \\
\qquad \textbf{for} it=1..numberOfIterations \\
\qquad \quad [Looking for a solution for each ant] \\
\qquad \qquad \textbf{for} each ant i \\
\qquad \qquad \qquad Place ant i at depot\\
\qquad \qquad \qquad i.shopsNotVisited $\leftarrow$ shops\\
\qquad \qquad \qquad i.solution  $\leftarrow \emptyset$ \\
\qquad \qquad \qquad i.solution $\leftarrow$ reset cost, time and demand \\
\qquad \qquad \qquad reset load \\
\qquad \qquad \textbf{while} there are ants i for which i.shopsNotVisited $\neq \emptyset$ \textbf{do} \\
\qquad \qquad \qquad \textbf{for} each ant i for which i.shopsNotVisited $\neq \emptyset$\\
\qquad \qquad \qquad \qquad nextShopToVisit $\leftarrow$ TransitionFunction()\\
\qquad \qquad \qquad \qquad Update cost\\
\qquad \qquad \qquad \qquad \textbf{if} nextShopToVisit == depot \textbf{then} reset time and load\\
\qquad \qquad \qquad \qquad \textbf{else} \\
\qquad \qquad \qquad \qquad \qquad Update time, demand and load\\
\qquad \qquad \qquad \qquad \qquad Update i.shopsNotVisited, i.solution\\
\qquad \qquad \qquad \qquad Place ant i at nextShopToVisit\\
\qquad \quad [Interchanges] \\
\qquad \qquad \textbf{for} each ant i \textbf{do} $\lambda$-interchanges in i.solution\\
\qquad \quad [Update pheromone matrix with local solution]\\
\qquad \qquad Find worstPathInIteration, bestPathInIteration\\
\qquad \qquad Reinforce bestPathInIteration in pheromone matrix\\
\qquad \qquad Reset worstPathInIteration in pheromone matrix\\
\qquad \quad [Update global solution]\\
\qquad \qquad \textbf{if} globalCost $>$ cost(bestPathInIteration)\\
\qquad \qquad \qquad \texttt{routes} $\leftarrow$ bestPathInIteration\\
\qquad \qquad \qquad   globalCost $\leftarrow$ cost(bestPathInIteration)\\
\qquad \quad [Update configuration]\\
\qquad \qquad Update algorithm parameters: $\rho$, p \\
\textbf{return} \texttt{routes}\\
\textbf{end algorithm;}  \\
%\hline
\end{tabular}
}}\caption{ACO - Main loop.} \label{alg:HybridACO-code}
\end{center}
\end{Algorithm}

\begin{Algorithm}[tb!]
\begin{center}\setlength{\tabcolsep}{2mm}
\renewcommand{\arraystretch}{1.2}
{\footnotesize \textsf{
\begin{tabular}{l}
 \multicolumn{1}{c}{}\\ %\hline
\textbf{Algorithm} Transition(shopsNotVisitedYet:List(shop),currentShop:shop,\\
\qquad remainingTime:float,remainingLoad:float):\texttt{nextShop} \\
\\
\qquad $j=currentShop$ \\
\qquad \textbf{if} ($q \leq p_t$) \\
\qquad \qquad $j=argmax_{shopsNotVisitedYet}
[\tau_{currentShop,j}]^\alpha[\eta_{currentShop,j}]^\beta[\mu_{currentShop,j}]^\gamma$\\
\qquad \textbf{else} \\
\qquad \qquad $j = randomFrom(shopsNotVisitedYet)$ \\
\qquad $time = TimeOf(currentShop,j) + TimeOf(j,Depot)$\\
\qquad \textbf{if} ($time \leq remainingTime$ AND
$DemandOf(j) \leq remainingLoad$ )\\
\qquad \qquad nextShop = j \\
\qquad \textbf{else} \\
\qquad \qquad nextShop = Depot \\
\textbf{return} \texttt{nextShop}\\
\textbf{end algorithm;}  \\
%\hline
\end{tabular}
}}\caption{\label{alg:transition-code} ACO - Transition function.}
\end{center}
\end{Algorithm}

\begin{Algorithm}[tb!]
\begin{center}\setlength{\tabcolsep}{2mm}
\renewcommand{\arraystretch}{1.2}
{\footnotesize \textsf{
\begin{tabular}{l}
 \multicolumn{1}{c}{}\\ %\hline
\textbf{Algorithm} TS \\
\qquad  [Initialisation] \\
\qquad \qquad currentSolution  $\leftarrow$  initialSolution \\
\qquad \qquad currentSolutionCost  $\leftarrow$  calculateCost(currentSolution)\\
\qquad  [Main loop] \\
\qquad \qquad \textbf{while} (iterations $<$ MAX-ITERATIONS) \\
\qquad \qquad \qquad bestNeighbour  $\leftarrow$  getBestNeighbour
(currentSolution, tabuList) \\
\qquad \qquad \qquad bestNeighbourCost  $\leftarrow$  calculateCost(bestNeighbour) \\
\qquad \qquad \qquad currentSolution  $\leftarrow$  bestNeighbour\\
\qquad \qquad \qquad currentSolutionCost  $\leftarrow$  bestNeighbourCost\\
\qquad \qquad \qquad \textbf{ if} (currentSolutionCost.isBetterThan(bestsolutionCost))\\
\qquad \qquad \qquad \qquad  bestSolution  $\leftarrow$  currentSolution\\
\qquad \qquad \qquad \qquad  bestSolutionCost  $\leftarrow$  currentSolutionCost \\
\qquad \qquad \qquad \qquad  iterations  $\leftarrow$  0 \\
\qquad \qquad \qquad \textbf{ else} \\
\qquad \qquad \qquad \qquad  iterations  $\leftarrow$  iterations + 1 \\
\qquad \qquad \qquad tabuList $\leftarrow$ updateTenure \\
\qquad    \textbf{return} \texttt{bestSolution}\\
\textbf{end algorithm;}  \\
%\hline
\end{tabular}
}}\caption{\label{alg:TS-code}Tabu Search (TS) algorithm - Main
loop.}
\end{center}
\end{Algorithm}

\begin{Algorithm}[tb!]
\begin{center}\setlength{\tabcolsep}{2mm}
\renewcommand{\arraystretch}{1.2}
{\footnotesize \textsf{
\begin{tabular}{l}
 \multicolumn{1}{c}{}\\ %\hline
\textbf{Algorithm} bestNeighbour\\
\qquad  [Initialisation] \\
\qquad \qquad moved $\leftarrow$ false\\
\qquad \qquad moves  $\leftarrow$  getAllMoves \\
\qquad \qquad theBestNeighbour $\leftarrow$ currentSolution\\
\qquad \qquad theBestNeighbourCost $\leftarrow \infty $ \\
\qquad \qquad neighbourCost $\leftarrow \infty $ \\
\qquad  [Main loop] \\
\qquad \qquad \textbf{for} i=1:moves.length\\
\qquad \qquad \qquad move $\leftarrow moves[i]$ \\
\qquad \qquad \qquad neighbour $\leftarrow$ currentSolution \\
\qquad \qquad \qquad neighbour $\leftarrow$ move.operateOn(neighbour) \\
\qquad \qquad \qquad neighbourCost $\leftarrow $ calculateCost(neighbour)\\
\qquad \qquad \qquad isTabu $\leftarrow $ isTabu(move)\\
\qquad \qquad \qquad [Aspiration criteria]\\
\qquad \qquad \qquad \textbf{if} (neighbourCost $<$ bestSolutionCost)\\
\qquad \qquad \qquad \qquad isTabu $\leftarrow$ false\\
\qquad \qquad \qquad \textbf{if} (neighbourCost $<$
theBestNeighbourCost \textbf{AND NOT} isTabu)\\
\qquad \qquad \qquad \qquad theBestNeighbour $\leftarrow$ neighbour
\\
\qquad \qquad \qquad \qquad theBestNeighbourCost $\leftarrow$
neighbourCost\\
\qquad \qquad \qquad \qquad bestNeighbourMove $\leftarrow$ move\\
\qquad \qquad \qquad \qquad moved $\leftarrow$ false\\
\qquad \qquad \textbf{end for}\\
\qquad  [Update tabu list] \\
\qquad \qquad\textbf{ if }moved == true\\
\qquad \qquad \qquad tabuList.addMove(bestNeighbourMove)\\
\qquad    \textbf{return} \texttt{theBestNeighbour}\\
\textbf{end algorithm;}  \\
%\hline
\end{tabular}
}}\caption{\label{alg:TS-bestneighbour-code}Tabu Search - Best
neighbour algorithm.}
\end{center}
\end{Algorithm}
\end{document}